\documentclass[11pt]{article}

\usepackage{amsmath}
\usepackage{graphicx}
\usepackage[numbers]{natbib}
\usepackage{url} 

\usepackage{inputenc}     
\usepackage[T1]{fontenc}    
\usepackage{hyperref}       
\usepackage{url}            
\usepackage{booktabs}       
\usepackage{amsfonts}       
\usepackage{nicefrac}       
\usepackage{microtype}      
\usepackage{times}
\usepackage{float}
\usepackage{algorithm}
\usepackage{makecell}
\usepackage{multirow}
\usepackage{hhline}
\usepackage{algorithmic}
\usepackage{amsmath,bm}
\usepackage{amsthm}
\usepackage{amssymb}
\usepackage[shortlabels]{enumitem}
\usepackage{graphicx}
\usepackage{color}
\usepackage{soul}
\usepackage{wrapfig}
\usepackage{subfigure}
\usepackage{dsfont}
\usepackage{pbox}
\usepackage{xr}

\newcommand{\ty}{\widetilde{y}}
\newcommand{\tby}{\widetilde{\bm{y}}}

\newcommand{\bT}{\bm{T}}

\newcommand{\bhbeta}{\widehat{\bbeta}}

\newcommand{\Norm}[1]{\left\Vert#1\right\Vert}

\newcommand{\Abs}[1]{\left\vert#1\right\vert}
\newcommand{\ind}[1]{\mathds{I}[#1]}
\newcommand{\Ind}[1]{\mathds{I}\left[#1\right]}

\def\singlespace{\def\baselinestretch{1}\@normalsize}

\newcommand{\bA}{{\mathbf A}}
\newcommand{\bB}{{\mathbf B}}

\newcommand{\bU}{{\mathbf U}}
\newcommand{\bV}{{\mathbf V}}

\newcommand{\bX}{{\mathbf X}}

\newcommand{\bu}{{\mathbf u}}
\newcommand{\bv}{{\mathbf v}}

\newcommand{\bx}{{\mathbf x}}
\newcommand{\by}{{\mathbf y}}

\newcommand{\bbeta}  {\boldsymbol{\beta}}

\newcommand{\bSigma}{\boldsymbol{\Sigma}}

\newcommand{\calD}{{\mathcal D}}

\newcommand{\calL}{{\mathcal L}}

\newcommand{\calN}{{\mathcal N}}

\newcommand{\calE}{{\mathcal E}}

\newcommand{\calG}{{\mathcal G}}

\def\6bullets{\bullet\bullet\bullet\bullet\bullet\bullet}

\DeclareMathAlphabet\EuScriptBF{U}{eus}{b}{n}

\newcommand{\Eb}{\mathbb{E}}

\newcommand{\I}{\mathbf{I}}

\newcommand{\p}{\mathbf{p}}

\renewcommand{\t}{\mathbf{t}}

\renewcommand{\u}{\mathbf{u}}

\renewcommand{\v}{\mathbf{v}}
\newcommand{\W}{\mathbf{W}}

\newcommand{\x}{\mathbf{x}}

\newcommand{\y}{\mathbf{y}}

\newcommand{\0}{\mathbf{0}}
\newcommand{\1}{\mathbf{1}}

\newcommand{\abs}[1]{\mbox{$\lvert #1 \rvert$}}
\newcommand{\norm}[1]{\mbox{$\left\lVert #1 \right\rVert$}}

\newtheorem{thm}{Theorem}
\newtheorem{cor}[thm]{Corollary}

\newtheorem{prop}[thm]{Proposition}

\newtheorem{rem}{Remark}

\newtheorem{assum}{Assumption}

\newcommand{\vbeta}{\boldsymbol \beta}

\newcommand{\vSigma}{\boldsymbol \Sigma}

\begin{document}
\title{Fast and Robust Sparsity Learning over Networks: A Decentralized Surrogate Median Regression Approach}
\author{Weidong Liu\thanks{Shanghai Jiao Tong University, Shanghai, 200240 China (e-mail: weidongl@sjtu.edu.cn).},
        Xiaojun Mao\thanks{Shanghai Jiao Tong University, Shanghai, 200240 China (e-mail: maoxj@sjtu.edu.cn).},
        and~Xin Zhang\thanks{Iowa State University, Ames, IA, 50011, USA (e-mail: xinzhang@iastate.edu).}
    \date{}
}

\maketitle

\begin{abstract}
Decentralized sparsity learning has attracted a significant amount of attention recently due to its rapidly growing applications. To obtain the robust and sparse estimators, a natural idea is to adopt the non-smooth median loss combined with a $\ell_1$ sparsity regularizer. However, most of the existing methods suffer from slow convergence performance caused by the {\em double} non-smooth objective. To accelerate the computation, in this paper, we proposed a decentralized surrogate median regression (deSMR) method for efficiently solving the decentralized sparsity learning problem. We show that our proposed algorithm enjoys a linear convergence rate with a simple implementation. We also investigate the statistical guarantee, and it shows that our proposed estimator achieves a near-oracle convergence rate without any restriction on the number of network nodes. Moreover, we establish the theoretical results for sparse support recovery. Thorough numerical experiments and real data study are provided to demonstrate the effectiveness of our method. 
\end{abstract}

\section{Introduction}\label{Section: introduction}

In recent years, decentralized machine learning (ML) has received growing research interest due to its advantages in system stability, data privacy, and computation efficiency \cite{Lian-Zhang-Zhang17,nedic2009distributed}. In contrast to the traditional centralized distributed architecture coordinated by a master machine, decentralized ML works with peer-to-peer networked systems, where workers can perform local computation and pass the message through the network links. 
The goal of decentralized ML is to learn a global ML model by having workers optimize their own models and share local model information with their neighbors.
So far, decentralized ML has achieved significant success in many scientific and engineering areas, including distributed sensing in wireless sensor networks\cite{ling2010decentralized,predd2006distributed,schizas2008consensus,zhao2002information}, multi-agent robotic systems\cite{cao2013overview,ren2007information,zhou2011multirobot}, smart grids\cite{giannakis2013monitoring,kekatos2013distributed} etc.

However, in spite of the increasing adoption in applications, the performances of most decentralized ML methods are not robust and are vulnerable to the following three aspects:
1) {\em Data Heterogeneity}.
Due to the lack of the global information aggregated by the central master, workers in decentralized network systems learn the model heavily relied on the local data and neighboring information.
Thus, data heterogeneity across workers often causes a so-called `consensus error' that reduces the model estimation efficiency \citep{nedic2009distributed,yuan2016convergence,li2020communication};
2) {\em Noise Contamination}.
In many real-world ML applications, training data are contaminated by heavy-tailed noises or outlier values.
Under this circumstance, adopting the mean square loss or Huber loss \citep{Huber73} in decentralized ML framework become less efficient or even infeasible \citep{Chen-Liu-Mao20};
3) {\em Cyber Attacks}. 
Without the central master, it is difficult for decentralized ML methods to identify the cyber attackers, which inject abnormal data into ML training.
Typically, these injected fake data will affect the training process and bias the estimated model.

In this paper, we aim to improve the estimation efficiency and robustness for decentralized ML. 
Specifically, we focus on the problem of decentralized sparsity learning on linear model. 
Consider a network with $m$ nodes, each of which holds a dataset $\calD_j= \{\x_{i},y_{i}\}_{i=1}^{n_j}$ for $j\in\{1,\ldots,m\}$, where $y\in\mathbb{R}$ is the observed response values, $\x=(x_1,\dots,x_p)^{\rm T} \in \mathbb{R}^p$ is a $p$-dimensional covariate vector. In the following we always use the notation $[t]$ as the set of indices $\{1,\ldots,t\}$ where the number $t$ can be changed to be the length of the set.
We suppose that the data follows a linear model:
\begin{align}\label{eq:model}
  y_{i} = \x_{i}^{\rm T}\vbeta^* +\epsilon_{i},~~\forall i \in [n_j], j \in [m],
\end{align}
where $\vbeta^*=(\beta_1,\dots,\beta_p)^{\rm T}$ is the globally common linear coefficient of dimension $p$, and $\epsilon_{i}$ is the measurement noise which is independent with covariate $\x_{i}$. Additionally, we consider that the coefficient $\vbeta^*$ follows a sparse structure with $s$ non-zero elements.
We denote the support of $\vbeta^*$ as $\mathcal{S}=\{1\leq i\leq p: \beta^{*}_{i}\neq 0\}$ and $|\mathcal{S}|=s$. 
Practically, we allow the data to be heterogeneous across nodes, i.e., the covariance structure $\bSigma_j=\Eb_{\x_i \in \calD_j}(\x_i\x_i^{\rm T})$ and distributions for measurement noise $\epsilon_{i}$ could be different with respect to node index $j$. 
Our primary interest is to estimate $\vbeta^*$ and recover its sparsity support by utilizing all the data across the network $\calD=\cup_{j=1}^{m} \calD_j$, where the size of $\calD$ is $N=\sum_{j=1}^{m}n_j$. For the ease of presentation, we assume that the data are evenly collected and each node has the sample size $|\calD_j|=n_j=n$ which implies $N=mn$.

There exist several decentralized sparsity learning methods \citep{Mateos-Bazerque-Giannakis10,chang2014multi,shi2015proximal,di2012sparse} that enjoy a fast linear convergence rate.
However, such convergence advantage benefits from the traditional least square loss, which is vulnerable to the abnormal value or outlier noise.
In this work, we allow the noise $\epsilon$ to be a heavy-tailed noise, of which the variance can be infinite. 
To mitigate the effect of heavy-tailed noise, a natural idea is to replace the mean square loss with the median loss, so that $\vbeta^*$ can be estimated by solving the following minimization problem,
\begin{align}\label{Eq: median_loss}
  \min_{\vbeta \in \mathbb{R}^p} \frac{1}{N}\sum_{j=1}^{m}\sum_{(\bx_i,y_i)\in\calD_j}\Abs{y_i-\bx_i^{\rm T}\vbeta},
\end{align}
where $\abs{\cdot}$ is for the absolute value.
To further pursue the sparsity structure of $\vbeta^*$, we regularize the problem~\eqref{Eq: median_loss} and define the sparse robust estimator for $\vbeta^*$ as
\begin{align}\label{Eq: median_loss_emp}
  \widehat{\vbeta} = \arg\min_{\vbeta \in \mathbb{R}^p} \frac{1}{N}\sum_{j=1}^{m}\sum_{(\bx_i,y_i)\in\calD_j}\Abs{y_i-\bx_i^{\rm T}\vbeta} + \lambda_N\abs{\vbeta}_1,
\end{align}
where $\abs{\vbeta}_1$ is the $\ell_1$-regularization term and $\lambda_N>0$ is the regularization parameter. 
However, due to the non-smooth property of both the median loss and the $\ell_1$-regularization term, it's a non-trivial task to optimize the above {\em double} non-smooth minimization problem~\eqref{Eq: median_loss_emp}, let alone solving it in a decentralized setting.
Although several decentralized subgradient methods were proposed to address the non-smooth median loss in network systems (e.g., \cite{Wang-Li17,wang2019distributed}), the estimation efficiency is not satisfactory because of their sublinear convergence rate.
The limitations of these existing works naturally arise the following question: {\em Could we develop a decentralized sparsity learning method for $\ell_1$-regularized median loss that can achieve a fast linear convergence rate?}

We give {\em affirmative} answer to the above question in this paper. We develop a Decentralized Surrogate Median Regression (deSMR) method and its key features are summarized in the followings:
\begin{itemize}
  \item To avoid the tremendous complexity caused by the non-smooth loss, we surrogate the median loss with a smooth least square loss with transformed responses.
  The idea of response transformation is inspired by Newton-Raphson updating. 
  We show that our method only needs a constant-order round of reformulation to achieve high accurate estimation;

  \item For collaboratively solving the reformulated sparsity-penalized least square loss, we derive a decentralized generalized ADMM algorithm, which enjoys a {\em linear} convergence.
  Our algorithm only requires nodes to find a closed-form solution for locally approximated subproblems and share local model parameters with their neighbors, and thus it is easy to implement.

  \item 
  By incorporating the algorithm convergence rate, we first establish the one-round node-wise statistical consistency rate for our proposed estimators in Theorem \ref{thm:betainf}.
  We show that our estimators recover the exact sparsity support under some mild conditions. Furthermore, we can iteratively improve the estimators over the networks by replacing the initial estimators.
  We show that after a constant number of iterations of loss surrogate and enough number of ADMM iterations, our estimators achieve a near-optimal rate of $\sqrt{s\log(N\vee p)/N}$ in Theorem \ref{thm:betainfV}. This rate matches with the optimal rate $\sqrt{s/N}$ obtained under the true support is known up to a logarithmic factor. Due to the space limitation, we leave the corresponding results of support recovery in the Supplementary Material.
\end{itemize}

\subsection{Related Work}\label{Section: Related Work}

In the literature, many methods have been developed for solving distributed sparse learning problem.
However, most of the literature focus on the centralized distributed setting \cite{Lee-Liu-Sun17,Battey-Fan-Liu18,Wang-Kolar-Srebro17,liu2019distributed,li2013distributed,zhou2018distributed}, of which the bottleneck is the central master:
for example, the limited bandwidth of the central master will undermine the training performance
and failure of the central master will directly ruin the system \cite{Lian-Zhang-Zhang17}. 
The authors of \cite{Mateos-Bazerque-Giannakis10} are the first few to investigate the sparsity learning problem in the peer-to-peer networked systems.
Specifically, they proposed two novel algorithms, distributed quadratic programming Lasso (DQP-Lasso) and distributed coordinate descent Lasso (DCD-Lasso), based on the alternating-direction method of multipliers \cite{gabay1976dual}.
But the two algorithms suffer the high computation complexity caused by updates for primal variable.
In \cite{chang2014multi}, the authors proposed Inexact Consensus-ADMM and showed that the proposed algorithm enjoys the linear convergence rate.
Furthermore, the authors of \cite{shi2015proximal,li2019decentralized} proposed the decentralized proximal gradient descent algorithms to bypass introducing the Lagrangian dual.
However, the above works mainly focus on the advances of the optimization convergence, while it remains unclear whether they could provide statistical guarantees on the consistency of model estimation and sparsity recovery. 
Additionally, these works study on the smooth loss for data fitting, e.g. the least square loss, which is not robust in term of heavy-tailed noise and outlier data.

To learn a robust sparse ML model, the quantile loss is widely adopted in statistics community \cite{Li-Zhu08,Belloni-Chernozhukov11,Wang-Wu-Li12,Zheng-Peng-He18}.
It is worthwhile noting that the median loss is a special case of the well known quantile loss $\rho_\tau(x)=x(\tau-\ind{x\leq 0})$ \citep{Koenker05} by setting the pre-specified quantile level $\tau=0.5$. Indeed our method can be easily extended to general quantile loss $\rho_\tau$ by shifting a constant to the response. For ease of presentation, we focus on analyzing the median loss. Motivated by the success of the parallel computing technique, a few recent works studied and developed distributed QR approaches.
\cite{Volgushev-Chao-Cheng19} proposed a divide-and-conquer approach to solving the QR problem in the centralized distributed system.
However, to maintain the statistical accuracy of the divide and conquer estimator, it requires that each local machine stores at least $o(N^{1/2}\log N)$ data, which would be infeasible when the memory of the local machine is limited.
And also, it requires that the local data are homogeneous, i.e., i.i.d. sample, which is usually violated in the network system.
To address these issues, \cite{Chen-Liu-Zhang19} proposed a linear estimator of QR (LEQR), by using a kernel smoothing technique.
The LEQR approach requires the master machine to store a subset of data and find an initial QR estimator, which will be broadcast to the local machines.
But LEQR focuses on the low-dimensional QR problem and cannot identify the sparsity of the QR regression coefficient.
\cite{Chen-Liu-Mao20} studied distributed $\ell_1$-regularized problem under the QR loss with sparsity recovery.
To deal with the quantile loss with the $\ell_1$ penalty, they transformed the responses and converted the distributed QR problem to a distributed ordinary linear regression.
Then the approximate Newton method was adopted to solve the distributed linear regression estimation problem. 
However, their proposed algorithm also requires that the local data are homogeneous. The above-mentioned approaches focus on the distributed system with a central master machine \footnote{Although \cite{Chen-Liu-Zhang19} proposed a tree-structured system to solve the QR problem in sensor network, their method still required a central root machine.}.

The most related to ours is \cite{Wang-Li17}, which is also addressing the QR problem in decentralized network systems. In \cite{Wang-Li17}, the authors proposed a decentralized subgradient algorithm. 
Their algorithm finds the global estimation via iteratively performing local subgradient updating and neighboring aggregations: the machines iteratively update the local estimates with the local subgradients, then share the local estimate with their network neighbors and calculate a weighted average based on the collected neighbors' estimates.
Following \cite{Wang-Li17}, a few other methods \cite{zhang2018distributed,wang2019distributed} were proposed to reduce the communication cost for solving QR problem in decentralized networks.
But these works adopt subgradient-based algorithms, which suffer the low convergence speed and cannot guarantee sparsity recovery.

The rest of this paper is organized as follows. Section \ref{Section: method} provides the decentralized median regression over the network and introduces the detailed algorithm. The theoretical results of the proposed estimators are established under mild regularity assumptions in Section \ref{sec:theory}. Section \ref{Section: experiment} reports the numerical experiments on the simulated and a real-life example is investigated in Section \ref{Section: realDataStudy}. A brief summary is provided in Section \ref{Section: conclusion}, and the technical details are given in Supplementary Materials.

\section{Methodology}\label{Section: method}

In this section, we first introduce the decentralized median regression in a networked distributed system and consider the equivalent consensus form. To handle the non-smooth loss, we propose a square loss transformation method. By constructing pseudo-response and encouraging sparsity, we transform it into an $\ell_1$-penalized least square regression problem, which becomes much computationally feasible. Based on this idea, we develop a generalized consensus ADMM algorithm in the networked system.

\subsection{Decentralized Median Regression with Sparsity}

In this section, we consider solving the above median regression problem \eqref{Eq: median_loss} in decentralized framework, of which there are a set of geographically dispersed computational nodes that form a network.
We represent a networked distributed computing system as an undirected connected network $\calG=(\calN,\calE)$, where $\calN$ and $\calE$ are the sets of nodes and edges, respectively, with $|\calN| = m$ as the node number. Denote $\calN_j$ to be the $j$-th node. In this connected network, we assume $N$ data samples are collected locally in these $m$ nodes $\calN_{1},\ldots,\calN_{m}$. As described in the introduction, each node $\calN_j$ collects its local data $\calD_j=\{(\bx_i,y_i):i\in[n]\}$. To be more practical, we allow the data to be heterogeneous across nodes.

The nodes have local computation capabilities and are only able to communicate with their neighbors via the edges in $\calE$. Its connectivity is modeled by an adjacency matrix $\W$ \cite{weisstein2007adjacency}. More precisely, $\W_{jk}\in \{0, 1\}$ denotes the connection between nodes $\calN_j$ and $\calN_k$. Nodes $\calN_j$ and $\calN_k$ have a pairwise communication link if and only if the corresponding weights $\W_{jk}=\W_{kj}$ are $1$; otherwise $0$.
For the network with no self-loops, the diagonal elements of $\W$ are $0$s.
In this work, we consider the network is connected and has no self-loops. 

Our goal is to have the computational nodes {\em distributively} and {\em collaboratively} solving the network-wide median regression problem as follows:
\begin{align}\label{Eq: general_problem}
\min_{\vbeta \in \mathbb{R}^p} \calL(\{(\bx_i,y_i)\!\in\!\calD \},\vbeta) \!=\! \min_{\vbeta \in \mathbb{R}^p} \frac{1}{m}\sum_{j=1}^{m} \calL(\{(\bx_i,y_i)\!\in\!\calD_j \},\vbeta),
\end{align}
where each local median loss function $\calL(\{(\bx_i,y_i)\in\calD_j \},\vbeta) \triangleq \frac{1}{n}\sum_{(\bx_i,y_i)\in\calD_j} \abs{y_i-\bx_i^{\rm T}\vbeta}+\lambda_N \abs{\vbeta}_1$ is only observable to node $j$. Note that in the loss function $\calL$, $\lambda_N$ is the parameter to adjust the sparsity of the estimated median regression coefficient, which is tuned to be universal over the network $\calG$. Because in decentralized network system, there is no central controller to maintain a common model parameter during the iterations \citep{nedic2009distributed,cao2013overview,sayed2014adaptation}.
Thus, to solve problem~\eqref{Eq: general_problem} in a decentralized fashion, we reformulate it in the following equivalent consensus form:
\begin{align}\label{Eq: consensus_problem}
&\min_{\bB \in \mathbb{R}^{mp}} \frac{1}{m}\sum_{j=1}^{m} \calL(\{(\bx_i,y_i)\in\calD_j \},\vbeta^{(j)}),\nonumber\\
&\centering ~~ \text{s.t.}~~ \vbeta^{(j)} = \vbeta^{(k)}, ~~ \forall (j,k) \in \calE,
\end{align}  
where $\bB\triangleq [\vbeta^{(1)\rm T},\cdots,\vbeta^{(m)\rm T}]^{\rm T}\in\mathbb{R}^{mp}$ is the concatenated version of the parameters $\vbeta^{(j)}$. Here $\vbeta^{(j)}$ is an introduced local parameter copy at node $j$. In problem~\eqref{Eq: consensus_problem}, the constraints enforce that the local parameter copy at each node is equal to those of its neighbors, hence the name ``consensus''. Clearly, the solution of problem~\eqref{Eq: general_problem} solves problem~\eqref{Eq: consensus_problem} and vice versa \citep{nedic2009distributed,shi2014linear,yuan2016convergence}. 

However, it is non-trivial to solve the above decentralized robust and sparse learning problem (\ref{Eq: consensus_problem}) because of the two factors: 1). The non-smooth median loss function makes decentralized gradient/Newton-based algorithms infeasible; 2). Composited with the sparsity $\ell_1$ regularizer, there is no close-form solution for the consensus loss function. To address the above two challenges, in the following, we will propose our method named Decentralized Surrogate Median Regression (deSMR) method to solve the above problem~\eqref{Eq: consensus_problem}.
Two key components are included in our method. We first perform a transformation on the response $y$ and surrogate the median loss with a least-square loss; Then, we adopt the generalized alternating direction method of multipliers (ADMM) framework to solve the decentralized penalized least square problem. 
We will show that our method enjoys a linear convergence rate and thus a low complexity.

\subsection{Square Loss Transformation for Outer Loop}\label{Sec. Loss Trans}

The penalized median loss in (\ref{Eq: general_problem}) follows a non-smooth loss plus non-smooth regularizer structure. To the best of our knowledge, the existing decentralized algorithms typically solve such {\em doubly} non-smooth loss functions by using subgradient-based algorithm \citep{Wang-Li17,zhang2018distributed,wang2019distributed}. 
However, one of the critical weaknesses of these works is that the subgradient-based algorithm suffers a slow {\em{sublinear}} convergence rate, which leads to the intensive computation and communication complexities in the decentralized network system. To avoid the tremendous complexity for solving the {\em doubly} non-smooth loss, in the following, we propose a square loss transformation to replace the median loss.

To faster the algorithm, we adopt the transformation methods in \cite{Chen-Liu-Zhang19} to reformulate the median regression problem \eqref{Eq: median_loss}. Inspired by the Newton-Raphson method, given the initial estimator $\vbeta_0$, \cite{Chen-Liu-Zhang19} proposed the new response variable $\ty$ as 
\begin{eqnarray}\label{eqn:newy}
  \ty=\bx^{\rm T}\vbeta_{0}-f^{-1}(0)(\mathds{I}[y\leq\bx^{\rm T}\vbeta_{0}]-1/2),
\end{eqnarray}
where $f(0)$ denotes the density of $\epsilon$ at $0$. With this transformation, it has been shown that the population version of the Newton-Raphson iteration to solve problem \eqref{Eq: median_loss} induces an updated solution $\vbeta_{1}$ as the least square solution of the problem $\Eb (\ty-\bx^{\rm T}\vbeta)^{2}$. 

In network node $\calN_j$ with the dataset $\calD_j$ from \eqref{eq:model}, suppose that we have the initial estimator $\bhbeta_{0}^{(j)}$ obtained by only using $\calD_j$ locally in the node $\calN_j$. We choose the corresponding density estimator of $f(0)$ to be 
\begin{eqnarray}\label{Eq: density estimation}
  \widehat{f}^{(j)}(0)=\frac{1}{nh^{(j)}}\sum_{(\bx_i,y_i)\in\calD_j}K_j\Big{(}\frac{y_i-\x_i^{\rm T}\widehat{\vbeta}_{0}^{(j)}}{h^{(j)}}\Big{)},
\end{eqnarray}
where $K_j(x)$ is a kernel function which satisfies the assumption (A3) (see Section \ref{sec:theory}) and $h^{(j)}\to 0$ are the bandwidths for each $j\in[m]$. The selection of bandwidths $\{h^{(j)}:j\in[m]\}$ will be discussed in our theoretical results (see Section \ref{sec:theory}). Moreover, for each $(\bx_i,y_i)\in\calD_j$, we construct the corresponding pseudo-response by
\begin{eqnarray}\label{Eq: pseudo-y}
  \tilde{y}_i=\x_i^{\rm T}\bhbeta_{0}^{(j)}-(\widehat{f}^{(j)}(0))^{-1}(\Ind{y_i\leq\x_i^{\rm T}\bhbeta_{0}^{(j)}}-1/2).
\end{eqnarray}

To further encourage the sparsity of the estimator, it is natural to consider the following $\ell_1$-regularized problem,
\begin{eqnarray}\label{eq:pool}
  \bhbeta_1=\arg\min\limits_{\vbeta\in R^{p}} \frac{1}{2N}\sum_{(\bx_i,y_i)\in\calD}(\widetilde{y}_i-\x_i^{\rm T}\vbeta)^{2}+\lambda_{N,0}\abs{\vbeta}_{1}.
\end{eqnarray}

Denote $\widetilde{\calL}(\{(\x_i,y_i)\in\calD_j \},\vbeta)=\frac{1}{2n}\sum_{(\x_i,y_i)\in\calD_j} (\tilde{y}_i-\bx_i^{\rm T}\vbeta)^2+\lambda_{N,0} \abs{\vbeta}_1$, in the decentralized framework, we focus on finding the minimizer of the following problem as surrogate estimator:
\begin{align}\label{Eq: transformed_general_problem}
\min_{\vbeta \in \mathbb{R}^p} \widetilde{\calL}(\{(\x_i,y_i)\!\in\!\calD \},\vbeta) \!=\!\min_{\vbeta \in \mathbb{R}^p} \!\frac{1}{m}\!\sum_{j=1}^{m} \widetilde{\calL}(\{(\x_i,y_i)\!\in\!\calD_j \},\vbeta).
\end{align}

Similarly as problem \eqref{Eq: consensus_problem}, we reformulate the above transformed general problem into the consensus form as:
\begin{align}\label{Eq: transformed_consensus_problem_1}
&\widehat{\bB}_1 = \arg\min_{\bB \in \mathbb{R}^{mp}} \frac{1}{2mn}\sum_{j=1}^{m}\Abs{\tby^{(j)}-\bX^{(j)}\vbeta^{(j)}}^2_2 + \lambda_{N,0}\Abs{\vbeta^{(j)}}_1,\nonumber\\
&\centering\text{s.t.}~~ \vbeta^{(j)} = \vbeta^{(k)}, ~~ \forall (j,k) \in \mathcal{E}.
\end{align}
where $\widehat{\bB}_1\!=\![\bhbeta_1^{(1)\rm T},\cdots,\bhbeta_1^{(m)\rm T}]^{\rm T} \!=\!\1_m \!\otimes\! (\frac{1}{m}\sum_{j=1}^{m}\widehat{\vbeta}^{(j)}) \in\mathbb{R}^{mp}$, $\tby^{(j)}=(\ty^{(j)}_1,\ldots,\ty^{(j)}_n)$ is the concatenated vector of dimension $p$, and $\bX^{(j)}=(\bx_1^{(j)},\dots,\bx_{n}^{(j)})$ is the concatenated covariate matrix of dimension $p\times n$ in node $\calN_j$.
 
It can be shown in Theorem \ref{thm:betainf} that the convergence rate of $\widehat{\bB}_1$ is not optimal under certain initial estimators. To further refine the estimator, we update the initial estimators iteratively by plugging-in $\widehat{\bB}_{v-1}$ in $v$-th iteration. Thus at $v$-th iteration, we consider the $\ell_1$-regularized problem to be
\begin{eqnarray}\label{eq:poolv}
\bhbeta_v=\arg\min\limits_{\vbeta\in R^{p}} \frac{1}{2N}\sum_{(\bx_i,y_i)\in\calD}(\widetilde{y}_{i,v}-\x_i^{\rm T}\vbeta)^{2}+\lambda_{N,v}\abs{\vbeta}_{1}.
\end{eqnarray}
And the corresponding consensus form of problem \eqref{eq:poolv} is:
\begin{align}\label{Eq: transformed_consensus_problem_t}
&\widehat{\bB}_v = \arg\min_{\bB \in \mathbb{R}^{mp}} \frac{1}{2mn}\sum_{j=1}^{m}\Abs{\tby_v^{(j)}-\bX^{(j)}\vbeta^{(j)}}^2_2 + \lambda_{N,v}\Abs{\vbeta^{(j)}}_1,\notag\\
& ~~ \text{s.t.}~~ \vbeta^{(j)} = \vbeta^{(k)}, ~~ \forall (j,k) \in \mathcal{E},
\end{align}
where $\widehat{\bB}_v=[\bhbeta_v^{\rm T},\cdots,\bhbeta_v^{\rm T}]^{\rm T}$. 


\subsection{Generalized Consensus ADMM for Inner Loop}\label{Sec2. inner_loop}

With the loss transformation proposed in Section~\ref{Sec. Loss Trans}, in the following, we focus on the derivation of detailed algorithm to solve the $v$-th iteration (\ref{Eq: transformed_consensus_problem_t}) consensus form problem. By introducing pseudo variables $\bT\triangleq[\t^{(jk)}]$, we can equivalently write the $v$-th iteration (\ref{Eq: transformed_consensus_problem_t}) as:
\begin{align}\label{Eq: ls_problem}
&\min_{\bB,\bT} \frac{1}{2mn}\sum_{j=1}^{m}\Abs{\tby_v^{(j)}-\bX^{(j)}\vbeta^{(j)}}^2_2 + \lambda_{N,v}\Abs{\vbeta^{(j)}}_1, \notag\\
& ~~ \text{s.t.}~~ \vbeta^{(j)} = \vbeta^{(k)} = \t^{(jk)}, ~~ \forall k\in\calN(j), j \in[m].
\end{align}
Due to the consensus constraint, a natural idea is to use the popular ADMM, which has been shown to be particularly efficient for solving linear constrained minimization problem.
Following classic ADMM \citep{gabay1976dual}, we can construct an augmented Lagrangian with penalty parameter $\tau >0$ as:
\begin{align}
&L_{v,\tau}(\bB,\bT,\bU,\bV) \!=\! \frac{1}{2mn}\!\sum_{j=1}^{m}\Abs{\tby_v^{(j)}-\bX^{(j)}\vbeta^{(j)}}^2_2 \!\!+\! \lambda_{N,v}\Abs{\vbeta^{(j)}}_1 \notag\\
&\!+\! \sum_{j=1}^{m}\sum_{k\in\calN(j)}\big(\langle \u^{(jk)}, \vbeta^{(j)} \!-\! \t^{(jk)} \rangle \!+\! \langle \v^{(jk)}, \vbeta^{(k)} \!-\! \t^{(jk)} \rangle \notag\\
&\!+\! \frac{\tau}{2} \Abs{\vbeta^{(j)}\! -\! \t^{(jk)}}_2^2 \!+\! \frac{\tau}{2}\Abs{\vbeta^{(k)}\! -\! \t^{(jk)}}_2^2 \big),
\end{align}
where $\bU\triangleq\{\u^{(jk)}\}_{j=1,k}^{m,\calN(j)}$ and $\bV\triangleq\{\v^{(jk)}\}_{j=1,k}^{m,\calN(j)}$ are the dual variables.
To find the solution for the above augmented Lagrangian, we define an axillary variable $\p^{(j)}_{v,t} =  \sum_{k\in\calN(j)} (\u^{(jk)}_{v,t}+\v^{(kj)}_{v,t})$ with $\p^{(j)}_0=\0$.
Then, (\ref{Eq: transformed_consensus_problem_t}) can be solved by recursively performing (\ref{Eq: update_p}) and (\ref{Eq: update_beta_3}):
\begin{subequations}
\label{eq:optim}
\begin{align}
&\p^{(j)}_{v,t+1 } \!= \!\p^{(j)}_{v,t}  \!+\! \tau \sum_{k\in\calN(j)} (\vbeta^{(j)}_{v,t} \!-\! \vbeta^{(k)}_{v,t})\label{Eq: update_p}.\\
&\vbeta^{(j)}_{v,t+1} \!=\! \arg\min\frac{1}{2mn}\Abs{\tby_v^{(j)}\!-\!\bX^{(j)}\vbeta^{(j)}}^2_2 \!+\! \lambda_{N,v}\Abs{\vbeta^{(j)}}_1\!\nonumber\\
&~~~~+\!\langle \p^{(j)}_{v,t+1}, \vbeta^{(j)}\rangle \!+\! \tau\!\!\! \sum_{k\in\calN(j)} \!\Abs{\vbeta^{(j)} \!-\! \frac{\vbeta^{(j)}_{v,t} \!+\! \vbeta^{(k)}_{v,t}}{2}}_2^2.
\label{Eq: update_beta_3}
\end{align}
\end{subequations}

Unfortunately, the minimization problem in (\ref{Eq: update_beta_3}) has no closed-form solution with a non-orthogonal matrix $\bX^{(j)}$, and it commonly requires multiple optimizing iterations to find an approximated minimizer \cite{Zhu17,Tao-Boley-Zhang16}.
To address this issue, we consider the generalized ADMM framework \cite{deng2016global,zhang2019distributed,gu2018admm,chang2014multi,Zhu17}, by adding a quadratic term $(\vbeta^{(j)}\!-\!\vbeta^{(j)}_{v,t})^{{\rm T}}(\rho_j\I \!-\! \frac{1}{mn}\bX^{(j){\rm T}}\bX^{(j)})(\vbeta^{(j)}\!-\!\vbeta^{(j)}_{v,t})$ to (\ref{Eq: update_beta_3}), and reach the close-form approximation of $\vbeta^{(j)}_{v,t+1}$:
\begin{align}\label{Eq: update_beta_4}
&\vbeta^{(j)}_{v,t+1} \!=\! \mathcal{S}_{2\lambda\omega_j}\big[\omega_j\big(\rho_j\vbeta_{v,t}^{(j)} \!- \!\frac{1}{mn}\bX^{(j){\rm T}}(\bX^{(j)}\vbeta_{v,t}^{(j)}\!-\!\tby^{(j)}_{v}) \notag\\
& ~~~~~~~~~~ - \!\p_{v,t+1}^{(j)} \!+\!\tau\!\!\!\sum_{k\in\calN(j)} (\vbeta_{v,t}^{(j)} \!+\! \vbeta_{v,t}^{(k)}) \big)\big],
\end{align}
where $\omega_j = 1/(\tau|\calN(j)| + \rho_j)$, $\mathcal{S}_{t}(\x)$ is the coordinate-wise soft-thresholding operator with $[\mathcal{S}_{t}(\x)]_i = (1-t/x_i)_+x_i$ and $(t)_+=t$ if $t>0$ , $0$ otherwise.
To summarize, we state our proposed decentralized algorithm for solving (\ref{Eq: general_problem}) in Algorithm~\ref{Algorithm: GT-QRE}.
Note that the inner loop updates in Algorithm~\ref{Algorithm: GT-QRE} can share the same rules with the Inexact Consensus ADMM \cite{chang2014multi} for $\ell_1$ penalized least square problem (\ref{Eq: ls_problem}).
We refer the readers to supplemental document for algorithm's derivation details.

\begin{algorithm}[H]
  \caption{Decentralized Surrogate Median Regression with Generalized Consensus ADMM.}\label{Algorithm: GT-QRE}.
  \begin{algorithmic} [1]
    \REQUIRE Local data $\calD_j \!=\! \{\bX^{(j)}, \by^{(j)} \}$ at node $\calN_j$, the number of outer iterations $V$ and inner iterations $T$, kernel functions $K_j(\cdot)$, bandwidths $\{h_{v}^{(j)}\}_{v=1,j=1}^{V,m}$, the tunning parameters $\lambda_{0,j}$ for initial estimator and universal tunning parameters $\lambda_{N,v}$ over the network.
    \STATE Locally compute the initial estimators $\widehat{\vbeta}_0^{(j)}$ at each node $\calN_j$:
    \begin{align}\label{Eq: local QR}
    \widehat{\vbeta}_0^{(j)} = \arg\min\limits_{\vbeta\in R^{p}} \frac{1}{n} \Abs{\y^{(j)} -\bX^{(j)\rm T}\vbeta}_1 + \lambda_{0,j}\abs{\vbeta}_1;
    \end{align}
    \FOR{$v = 0,\cdots, V$}
    \STATE Construct pseudo responses $\tby_{v}^{(j)}$ as in (\ref{Eq: pseudo-y}) at each node $\calN_j$ locally;
    \STATE Set $\vbeta_{v,0}^{(j)}= \widehat{\vbeta}_{v}^{(j)}$ and $\p_{v,0}^{(j)} = \0$;
    \FOR{$t = 1,\cdots, T$}
    \STATE Communicate local parameter $\vbeta_{v,t}^{(j)}$ with neighboring nodes;
    \STATE Update $\p_{v,t}^{(j)}$ and $\vbeta_{v,t}^{(j)}$ with (\ref{Eq: update_p}) and (\ref{Eq: update_beta_4}), respectively;
    \ENDFOR
    \STATE Set $\widehat{\vbeta}_{v+1}^{(j)} = \vbeta_{v,T+1}^{(j)}$;
    \ENDFOR
  \end{algorithmic}
\end{algorithm}

\begin{rem}\label{rem:penalty}
  Thanks to the generality of the outer loops' transformation and inner loops' ADMM updates, our algorithm can be easily extended to handle other penalties such as $\ell_0$ or SCAD\citep{Fan-Li01}, beyond the $\ell_1$ penalty. 
  More specifically, at each outer loop, we can perform the same transformation of the response variable as \eqref{Eq: pseudo-y}, and replace the regularizer in \eqref{eq:pool} and \eqref{Eq: transformed_consensus_problem_t} by the corresponding penalties. 
  Similarly, at the inner loop, we can directly change the penalty function when constructing the augmented Lagrangian.
\end{rem}

\section{Theoretical Results}\label{sec:theory}

For a vector $\bv=(v_{1},\dots,v_{p})^{\rm T}$, define the $\ell_1$ norm $\abs{\bv}_{1}=\sum_{i=1}^{p}\abs{v_{i}}$ and the $\ell_2$ norm $\abs{\bv}_{2}=\sqrt{\sum_{i=1}^{p}v_{i}^{2}}$. For a matrix $\bA=(a_{ij})\in\mathbb{R}^{m\times n}$, define $\abs{\bA}_{\infty}=\max_{1\le i\le m,1\le j \le n}\abs{a_{ij}}$ as the infinity norm, $\norm{\bA}_{F}=\sqrt{\sum_{i=1}^{m}\sum_{j=1}^{n}a_{ij}^{2}}$ as the Frobenius norm, $\norm{\bA}=\max_{\abs{\bu}_2=1} \abs{\bA \bu}_2=\sigma_{\max}(\bA)$ as the spectral norm (the largest singular value), and $\norm{\bA}_{\infty}=\max_{1\le i\le m}\sum_{j=1}^{n}\abs{a_{ij}}$. Define $\lambda_{\text{max}}(\bA)$ and $\lambda_{\text{min}}(\bA)$ to be the largest and smallest eigenvalues of $\bA$ respectively. Denote two subsets of indices $I=\{i_1,\ldots,i_r\}\subseteq[m]$ and $J = \{j_1,\ldots,j_q\}\subseteq [n]$, we use $\bA_{I\times J}$ to denote the $r$ by $q$ submatrix given by $(a_{i_sj_t})$. For two sequences $a_n$ and $b_n$, $a_n \asymp b_n$ if and only if both $a_n = O(b_n) $ and $b_n = O(a_n) $ hold simultaneously. 

\subsection{Technique Assumptions}\label{subsec:assum}
In this section, the technical assumptions needed for our analysis are given as follows. We first provide the assumptions required to achieve linear convergence rate of our proposed algorithm.
\begin{assum}\label{ass:network}
  The peer-to-peer networked system $\mathcal{G}$ is connected. 
\end{assum}
\vspace{-.2in}
\begin{assum}\label{ass:subG}
  For all the covariate $\bx$, they satisfy the sub-Gaussian condition for some constants $t>0$ and $C>0$, $$\sup\nolimits_{\abs{\bm\theta}_{2}=1}\Eb\exp(t(\bm\theta^{\rm T}\bx)^2)\le C.$$
\end{assum}
\vspace{-.2in}
\begin{assum}\label{ass:covariance}
  The dimension $p$ over the local sample size $n$ is $p/n\le\tau$ where $\tau\in(0,1)$. 
  For each $j\in[m]$, suppose that $\bSigma_j=\Eb(n^{-1}\bX^{(j)}\bX^{(j){\rm T}})$ satisfies $c_{0,j}^{-1}\le\lambda_{\text{min}}(\bSigma_j)\le\lambda_{\text{max}}(\bSigma_j)\le c_{0,j}$ for some constants $c_{0,j}>0$.
\end{assum}
  Assumption \ref{ass:network} is a common assumption in network consensus optimization. It indicates that there is no isolated nodes or groups of nodes in the network. Thus, all the nodes could reach a consensus status. Assumption \ref{ass:subG} is a regular sub-Gaussian assumption on the distribution of the covariates $\bX$ over the network $\calG$. Assumption \ref{ass:covariance} is a standard assumption, which implies that the largest eigenvalue of empirical covariance matrix $n^{-1}\bX^{(j)}\bX^{(j){\rm T}}$ is bounded with high probability goes to one together with Assumption \ref{ass:subG}. According to \cite{Yaskov14}, the dimension restriction is to ensure the empirical smallest eigenvalue of each sample covariance matrices $n^{-1}\bX^{(j)}\bX^{(j){\rm T}}$ are bounded away from zero with high probability tending to one. Thus the optimization problem is strongly convex.

To further obtain the statistical convergence rates at each iterations, we assume the following regular assumptions.
  \begin{assum}\label{ass:irres}
    For each $j\in[m]$, there exists some incoherence parameters $0<\alpha_j<1$ such that
    \begin{equation}\label{eqn:irres}
    \Norm{\bm{\Sigma}_{j,S^{c}\times S}\bm{\Sigma}_{j,S\times S}^{-1}}_{\infty}\le 1-\alpha_j.
    \end{equation}
  \end{assum}
\vspace{-.2in}
  \begin{assum}\label{ass:dimen}
    The dimension $p$ satisfies $p=O(N^{\nu})$ for some $\nu>0$. The local sample size $n$ on each node satisfies $n\geq N^{c}$ for some $0<c<1$, and the sparsity level $s$ satisfies $s=O(n^{r})$ for some $0<r<1/3$.
  \end{assum}
\vspace{-.2in}
  \begin{assum}\label{ass:f0} 
    Over the network $\calG$, the noises enjoy universal density function $f(\cdot)$ which is bounded and Lipschitz continuous (i.e., $\abs{f(\bx)-f(\by)}\le C_{L}\abs{\bx-\by}_2$ for any $\bx,\by\in\mathbb{R}^p$ and some constant $C_{L}>0$). Moreover, we assume $f(0)$ is bounded away from $0$.
  \end{assum}
\vspace{-.2in}
  \begin{assum}\label{ass:kernel} 
    For each $j\in[m]$, assume that the kernel function $K_j(\cdot)$ is integrable with $\int_{-\infty}^\infty K_j(u)\mathrm{d}u = 1$. Moreover, assume that $K_j(\cdot)$  satisfies $K_j(u)=0$ if $|u|\ge 1$. Further, assume $K_j(\cdot)$ is differentiable and its derivative $K_j'(\cdot)$ is bounded. 
  \end{assum}
\vspace{-.2in}
  \begin{assum}\label{ass:initial}
    For each $j\in[m]$, the initial estimators $\bhbeta_{0}^{(j)}$ satisfies $\abs{\bhbeta_{0}^{(j)}-\bbeta^{*}}_{2}=O_{\mathbb{P}}(\sqrt{s(\log N)/n})$. Furthermore, assume that $\Pr(\text{supp}(\bhbeta_{0}^{(j)})\subseteq S)\rightarrow 1$.
  \end{assum}
    Together with Assumption \ref{ass:covariance}, Assumption \ref{ass:irres} is commonly considered as irrepresentable condition in high-dimensional statistics literature \citep{Zhao-Yu06,Wainwright09,Buhlmann-Van-De-Geer11,Hastie-Tibshirani-Wainwright15} to establish results on support recovery. Assumption \ref{ass:dimen} is required on dimension $p$, local sample size $n$ on each node and sparsity level $s$ of the true coefficient. This assumption makes sure that our estimator achieves the near-oracle convergence rate only using a finite number of outer loop iterations with a sufficient number of inner loop iterations. Assumption \ref{ass:f0} is a regular assumption on the smoothness of the density function $f(\cdot)$. Note that we assume the noises have a universal density function $f(\cdot)$ instead of different on each node. Assumption \ref{ass:kernel} is standard on the kernel functions $K_j(\cdot) $. Although we allow using different kernel functions on each node, there is no technical difficulty obtaining Lemma of the Supplement Materials. For the ease of computation, we adopted one same kernel function $K(\cdot)$ during the implementation of our proposed deSMR. 

    Assumption \ref{ass:initial} is an assumption on the convergence rate and support recovery of the initial estimators at each node. Note that in the proposed Algorithm \ref{Algorithm: GT-QRE}, the initial estimators $\bhbeta_{0}^{(j)}$ is obtained as the solution to the high-dimensional median regression problem using data locally on each node $\calN_j$. It has been shown in \cite{Fan-Fan-Barut14} that each initial estimator $\bhbeta_{0}^{(j)}$ obtained by \eqref{Eq: local QR} fulfills Assumption \ref{ass:initial} under Assumptions \ref{ass:covariance}-\ref{ass:subG} and certain regularity conditions. Moreover, in $v$-th iteration in outer loop for the estimators $\bhbeta_{v,T+1}^{(j)}$ we obtained at each node $\calN_j$, we shown that Assumption \ref{ass:initial} still holds. Note that we use $\log(N)$ in the following convergence rates for notation simplicity due to the observation that $\log(\max(N,p))=C_1 \log(N)$ for some constant $C_1>0$ under the assumption $p=O(N^\nu)$.
    
\subsection{Convergence of Generalized Consensus ADMM}

In this section, we establish the convergence properties of our proposed ADMM algorithm in deSMR. We show that with certain assumptions and proper choice of the step lengths $\rho_j$, the algorithm yields a linear convergence rate in the connected network.

\begin{prop}[Linear Convergence]\label{prop:linear}
  Under Assumptions \ref{ass:network}-\ref{ass:covariance}, by setting the step lengths 
  $\rho_j > \lambda_{\text{max}}(n^{-1}\bX^{(j)}\bX^{(j){\rm T}})$ 
  for each $j\in[m]$, at $v$-th iteration where $v\in[V]$, it holds that 
  \begin{align}
    \|\bB_{v,T+1} - \widehat{\bB}_{v}\|^2_F = O_{\mathbb{P}}(\gamma^{T}),
  \end{align}
  where $\bB_{v,T+1} = [\vbeta_{v,T+1}^{(1){\rm T}},\cdots,\vbeta_{v,T+1}^{(m){\rm T}}]^{{\rm T}}$ and $\gamma \in (0,1)$.
\end{prop}

The result in Proposition \ref{prop:linear} shows that our inner loop algorithm can solve the reformulated $\ell_1$ penalized least square loss at a linear convergence speed, where the convergence factor $\gamma$ depends on the network topology $\W$ and singularity of the covariance matrices $\{\bSigma_j\}_{j=1}^{m}$.
Note that the step lengths selection is only relied on the maximum eigenvalue of the local covariance matrix, which makes the implementation simple. 
Due to the limited space, the proof of Proposition \ref{prop:linear} is relegated to the supplementary materials.

\subsection{Statistical Properties of Estimator $\bB_{v,T+1}$}

In this section we provide the theoretical results for the solutions $\bB_{v,T+1}$ obtained by our proposed decentralized method at $v$-th iteration. Recall that the support of true coefficient $\vbeta^*$ to be $S=\{1\leq i\leq p: \beta^{*}_{i}\neq 0\}$. 

Let $\{a_{N,v,j}\}$ be the convergence rate of the initial estimators $\{\bhbeta_{v,T+1}^{(j)}\}$ at $v$-th iteration. By Assumption \ref{ass:initial} we can assume that $a_{N,0,j}=\sqrt{s(\log N)/n}$ for each $j\in[m]$. We first provide the convergence rate for $\bB_{1,T+1}=\{\bbeta_{1,T+1}^{(1)\rm T},\cdots,\bbeta_{1,T+1}^{(m)\rm T}\}$ after one iteration in outer loop.

\begin{thm}\label{thm:betainf}
  For each $j\in[m]$, let $\abs{\bhbeta_{0}^{(j)}-\bbeta^{*}}_{2}=O_{\mathbb{P}}(a_{N,0,j})$ and choose the bandwidth $h^{(j)}\asymp a_{N,0,j}$, take 
  $$\lambda_{N,0}=C_{0}\left(\sqrt{\frac{\log N}{N}}+\max_j\{a_{N,0,j}\}\sqrt{\frac{s\log N}{n}}\right),$$ with $C_{0}$ being a sufficiently large constant in problem \ref{eq:pool}. Assume that Assumptions \eqref{ass:covariance}-\eqref{ass:initial} hold, we have for each $j\in[m]$ and $\gamma \in (0,1)$,
  \begin{align}\label{eqn:betainf}
  \Abs{\!\bbeta_{1,T+1}^{(j)}\!\!-\!\bbeta^{*}\!}_{2}\!\!\!\!\!=\!O_{\mathbb{P}}\Big(\!\!\sqrt{\frac{s\log N}{N}}\!+\!\max_j\{a_{N,0,j}\}\sqrt{\frac{s^2\log N}{n}}\!+\!\gamma^{T}\!\Big).
  \end{align}
\end{thm}

Compared with the initial estimators $\bhbeta_{0}^{(j)}$, our estimators $\bbeta_{1,T+1}^{(j)}$ obtained by deSMR improve the convergence rate to be $\max\{\sqrt{s(\log N)/N},\max_j\{a_{N,0,j}\}\sqrt{s^{2}(\log N)/n},\gamma^{T}\}$. Note that the first two terms comes from the order of $\abs{\bhbeta_{1}-\bbeta^{*}}_{2}$ where $\bhbeta_{1}$ is defined in problem \ref{eq:pool}. Note that this is indeed a pooled version least-square problem which put all the data collection together. It has been well studied \cite{Van-de-Geer08} that the regular least-square estimator can achieve convergence rate of $\sqrt{s(\log N)/N}$. Our second term is due to the construction of new response variables $\widetilde{y}_{i}$ to against the existence of heavy tail. Note that the third term denotes the order of $\abs{\bbeta_{1,T+1}^{(j)}-\bhbeta_{1}}_{2}$ which comes from the proposed deSMR algorithm. Under Assumptions \ref{ass:dimen} and \ref{ass:initial}, by allowing sufficient large iteration number of inner loop $T\ge \log(s^{2}(\log N)/n)/\log(\gamma)$, we have $\sqrt{s^{2}(\log N)/n} = o(1)$ and thus $a_{N,1,j}=o(a_{N,0,j})$. More specifically, we can further conclude that $T$ only need to be constant order according to the following Corollary.
\begin{cor}\label{cor:betainf}
  Under the Assumptions in Theorem \ref{thm:betainf}, when $N=O(\exp(n/s^2))$, we further have the iteration number of inner loop $T\ge \log(s^{2}(\log N)/n)/\log(\gamma)$ only need to be constant order.
\end{cor}

By recursive updating the initial estimators at each node $\calN_j$, we refine the convergence rate of the multi-iteration estimator $\bbeta_{v,T+1}^{(j)}$ iteratively. Before that, we provide results on support recovery of the proposed estimators $\bbeta_{1,T+1}^{(j)}$ to ensure Assumption \ref{ass:initial} holds for $v$-th outer loop iteration. Let $\bbeta_{1,T+1}^{(j)}=(\beta_{1,T+1,1}^{(j)},\beta_{1,T+1,2}^{(j)},\ldots,\beta_{1,T+1,p}^{(j)})^{\rm T}$ and denote $\widehat{S}_{1}^{(j)}=\{s:\beta_{1,T+1,s}^{(j)}\neq0\}$.

\begin{thm}\label{thm:support}
  With the same assumptions in Theorem \ref{thm:betainf}. For each $j\in[m]$, we have $\widehat{S}_{1}^{(j)}\subseteq S$ with probability tending to one. In addition, suppose that for some sufficiently large constant $C>0$,
  \begin{align}\label{eqn:sigcon}
  \underset{s\in S}{\min}\Abs{\beta^{*}_{s}}\ge& C\Norm{\left(\frac{1}{m}\sum_{j=1}^{m}\bSigma_{j,S\times S}\right)^{-1}}_{\infty}\times\\
  &\left(\sqrt{\frac{\log N}{N}}+\max_j\{a_{N,0,j}\}\sqrt{\frac{s\log N}{n}}+\gamma^{T}\right).\nonumber
  \end{align}
  Then we have $\widehat{S}_{1}^{(j)} = S$ with probability tending to one for each $j\in[m]$.
\end{thm}

Note that the results $\widehat{S}_{1}^{(j)}\subseteq S$ with probability tending to one for each $j\in[m]$ follow directly from Theorem \ref{thm:betainfV}. As for the ``beta-min'' condition, it matches with the convergence rate of $\bbeta_{1,T+1}^{(j)}$. In the next we show the improvement of convergence rates which achieve a near-oracle rate after a constant iteration number of outer loops. As for the similar support recovery statement, we leave it to Supplementary Material. For the $v$-th initials estimators $\bhbeta_{v}^{(j)}=\bbeta_{v,T+1}^{(j)}$, we define the 
\begin{eqnarray}\label{eq:a}
a_{N,v,j}=\sqrt{\frac{s \log N}{N}}+s^{(2v+1)/2}\left(\frac{\log N}{n}\right)^{(v+1)/2}+\gamma^{T},
\end{eqnarray}
for any $1\leq v\leq V$. Note that $a_{N,v,j}$ also include the optimization error induced by deSMR.

\begin{thm}\label{thm:betainfV}
  For each $j\in[m]$, assume that the initial estimators $\bhbeta_{v-1}^{(j)}=\bbeta_{v-1,T+1}^{(j)}$ satisfies $\abs{\bhbeta_{v-1}^{(j)}-\bbeta^{*}}_{2}=O_{\mathbb{P}}(a_{N,v-1,j})$. Let $h_{v}^{(j)}\asymp a_{N,v-1,j}$ for $v=2,\dots,V$, and take
  \begin{equation}\label{eq:lambda}
  \begin{aligned}
  \lambda_{N,v}=C_{0}\left(\sqrt{\frac{\log N}{N}}+\max_j\{a_{N,v-1,j}\}\sqrt{\frac{s\log N}{n}}\right),
  \end{aligned}
  \end{equation} with $C_{0}$ being a sufficiently large constant. Assume that Assumptions \eqref{ass:covariance}-\eqref{ass:initial} hold, we have
  \begin{equation}\label{eq:bt}
  \Abs{\bbeta_{v,T+1}^{(j)}\!-\!\bbeta^{*}}_{2}\!\!=\!O_{\mathbb{P}}\Big(\!\!\sqrt{\frac{s \log N}{N}}\!+\!s^{\frac{2v+1}{2}}\left(\frac{\log N}{n}\right)^{\frac{2v+1}{2}}\!\!\!+\!\gamma^{T}\Big).
  \end{equation}  
\end{thm}
When the outer loop iteration number $v$ is sufficiently large, i.e.,
\begin{equation}\label{eq:t}
v\geq \frac{\log (N/n)}{\log (\min\{c_{j}\}n/(s^2\log N))},\quad \text{for some }c_{j}>0,
\end{equation}
we have the second term in \eqref{eq:bt} dominated by the first term. Under Assumption \ref{ass:dimen}, the right hand side is bounded by a constant. By allowing iteration number of inner loop to be sufficient large such that $T\ge \log(s(\log N)/N)/2\log(\gamma)$, the convergence rate in \eqref{eq:bt} becomes $\abs{\bbeta_{v,T+1}^{(j)}-\bbeta^{*}}_{2}=O_{P}(\sqrt{s(\log N)/N})$. We note that under the Assumption \ref{ass:dimen}, the iteration number of inner loop $T$ required here is the same order as $\log(s^2(\log N)/n)/\log(\gamma)$ in the first outer iteration. For the ease of implementation, we use a universal large enough iteration number of inner loop $T$ in Algorithm \ref{Algorithm: GT-QRE}. Moreover, our optimal rate $\sqrt{s(\log N)/N}$ nearly matches the oracle convergence rate $\sqrt{s/N}$ (up to a logarithmic factor) with the prior knowledge of the support $S$. 

\section{Experimental Evaluation}\label{Section: experiment}

In this section, we empirically examine the computational and statistical performance of our proposed deSMR and decentralized robust median estimator.

\subsection{Simulation Setup}\label{Sec3. Setup}

We consider a decentralized network system with $m$ workers generated by Erd$\ddot{\text{o}}$s-R$\grave{\text{e}}$nyi graph with the connection probability $p_c$. 
At each node, data $\calD_j=\{\x_i,y_i\}_{i=1}^{n}$ follows the linear model $y_i = \x_i^{\rm T}\vbeta^* +\epsilon_i$,
where $\x_i=(x_{i,1},\dots,x_{i,p})^{\rm T} \in \mathbb{R}^p$ is a $p$-dimensional covariate vector, $\vbeta^*=(\beta_1,\dots,\beta_p)^{\rm T}$ is the true regression coefficient, and $\epsilon$ is the noise.
We generate the covariate $\x_i$ by $i.i.d.$ sampling from a multivariate normal distribution $N(0,\vSigma)$. The covariance matrix $\vSigma$ is constructed by $\vSigma_{ij} = \sigma^2 \rho ^{|i-j|}$ for $1\leq i,j\leq p$, where $\sigma^2$ and $\rho>0$. 
The true coefficient is set as $\vbeta^* = (1,2,\ldots,10,0,0\ldots,0) \in \mathbb{R}^p$ so that the sparsity level is $s = 10$. 
Without specification, we set $m=10$, $p_c=0.3$, $\sigma^2=1$, and $\rho = 0.1$.
Each simulation result is based on 100 independent repetitions.

To obtain the initial estimator, each node solves the local $\ell_1$-regularized median regression problem (\ref{Eq: local QR}) with the function \texttt{LASSO.fit} in R package \texttt{rqPen}.
At each outer loop iteration, we apply Bayesian information criterion (BIC) to choose $\lambda_{N,v}$.
For density estimation in (\ref{Eq: density estimation}), we use a bi-weight kernel function
\begin{align}
K(x) \!= \!\begin{cases}
0, & \text{if} \quad \!x\!\leq\! -1,\\
-\!\frac{315}{64}x^6\!+\!\frac{735}{64}x^4\!-\!\frac{525}{64}x^2\!+\!\frac{105}{64}, & \text{if} \quad \!-1\!\leq\! x\!\leq 1,\\
0, & \text{if} \quad \!x \!\geq\!1.
\end{cases}\notag
\end{align}
It is easy to verify that $K(\cdot)$ satisfies the condition (C3). 
Additionally, we choose the bandwidth as $h_v = \sqrt{\frac{s \log n}{n}}+s^{-1/2}\left(c_0\frac{s^2\log n}{m}\right)^{(v+1)/2}$.
Note that the constant $c_0$ is used to ensure that $\frac{s^2\log n}{m}<1$, and we set $c_0=0.013$ in the following experiments.
 
\subsection{Effect of Iteration Number}\label{Sec3. Effect of Iteration}

Our first numerical study focuses on the computational performance of our deSMR method.
Recall that our deSMR method has a double structure: 
In the outer loops, we perform a transformation on the response and convert the median loss to a least-square loss; 
Then in the inner loops, we adopt the decentralized ADMM algorithm to solve the penalized least square loss iteratively. 
Thus, in this part, we aim to study the impact of both the outer and inner loop iteration rounds on the estimation efficiency.

\begin{figure*}[!ht]
\vspace{-.1in}
    \centering
    \subfigure[Gaussian Noise]{
        \includegraphics[width=0.31\textwidth]{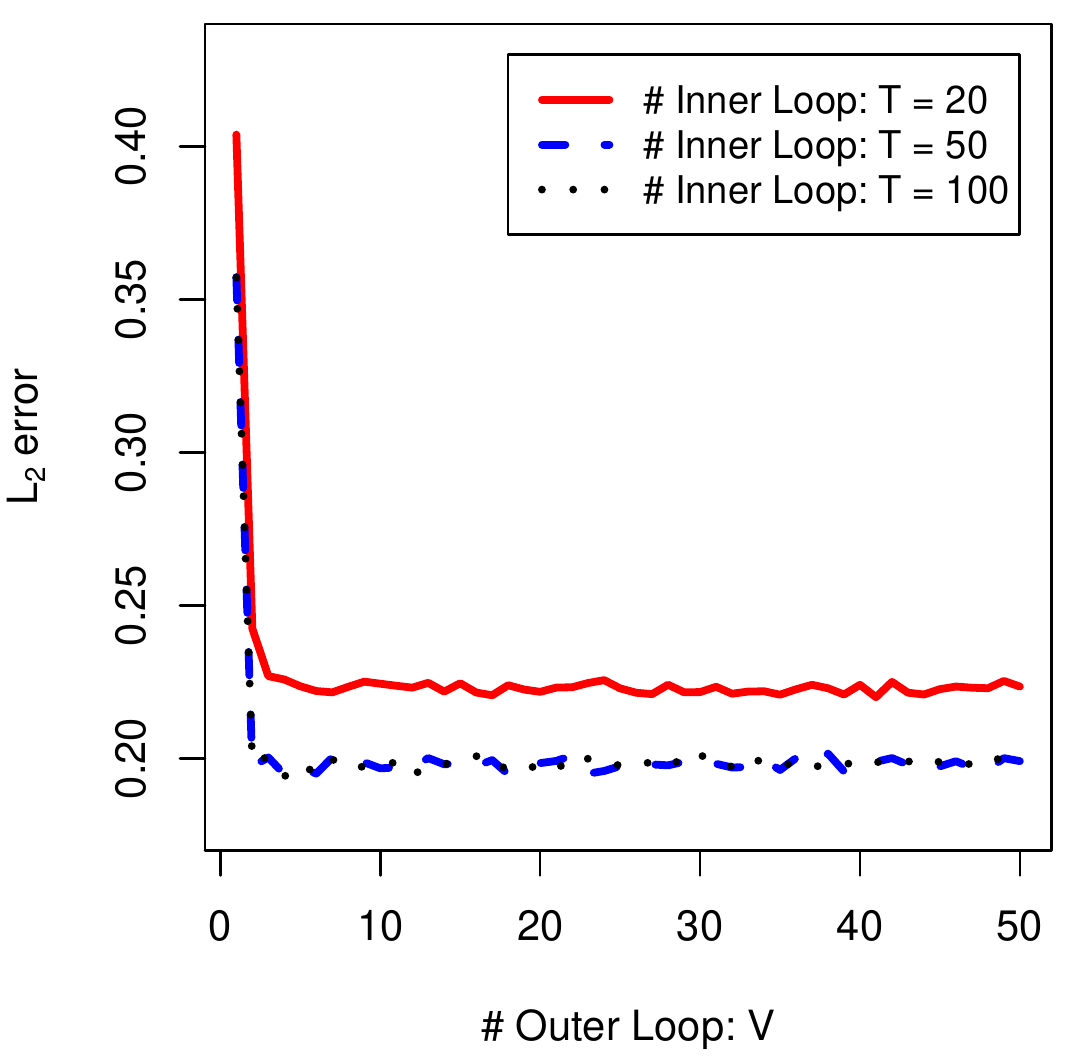}
        \label{fig:simu1_gaussian}
    }
    \subfigure[Exponential Noise]{
        \includegraphics[width=0.31\textwidth]{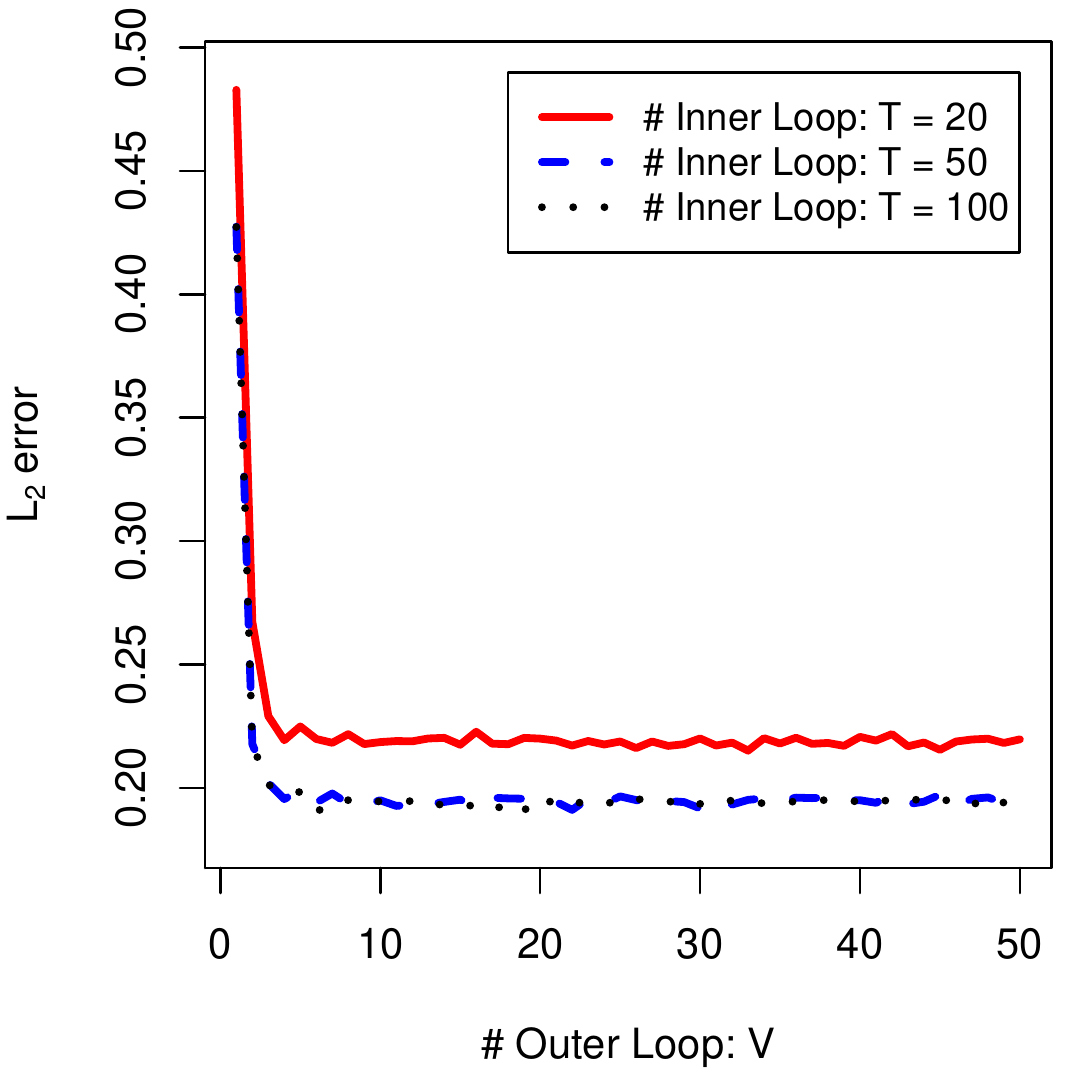}
        \label{fig:simu1_exp}
    }
    \subfigure[Cauchy Noise]{
        \includegraphics[width=0.31\textwidth]{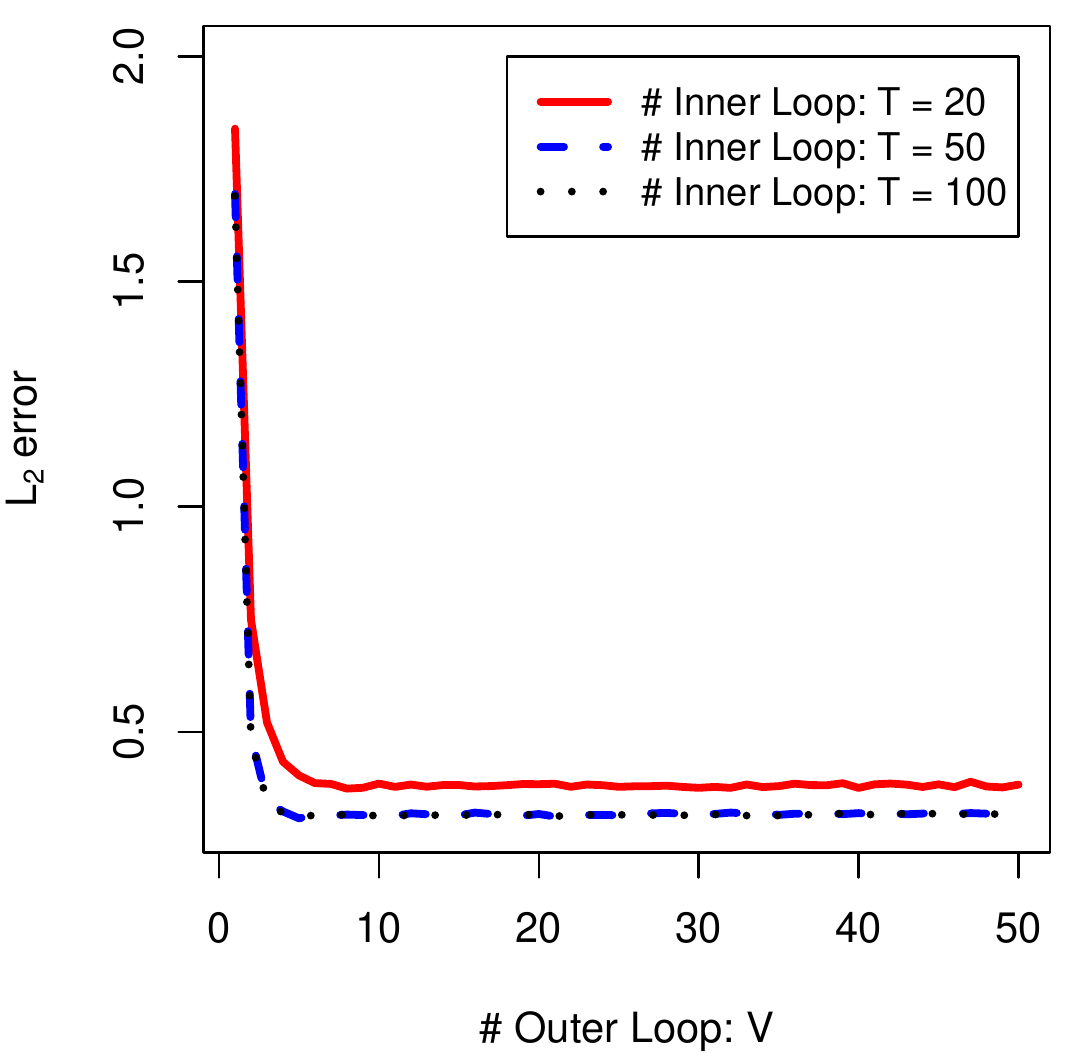}
        \label{fig:simu1_cauchy}
    }
    \subfigure[Gaussian Noise]{
        \includegraphics[width=0.31\textwidth]{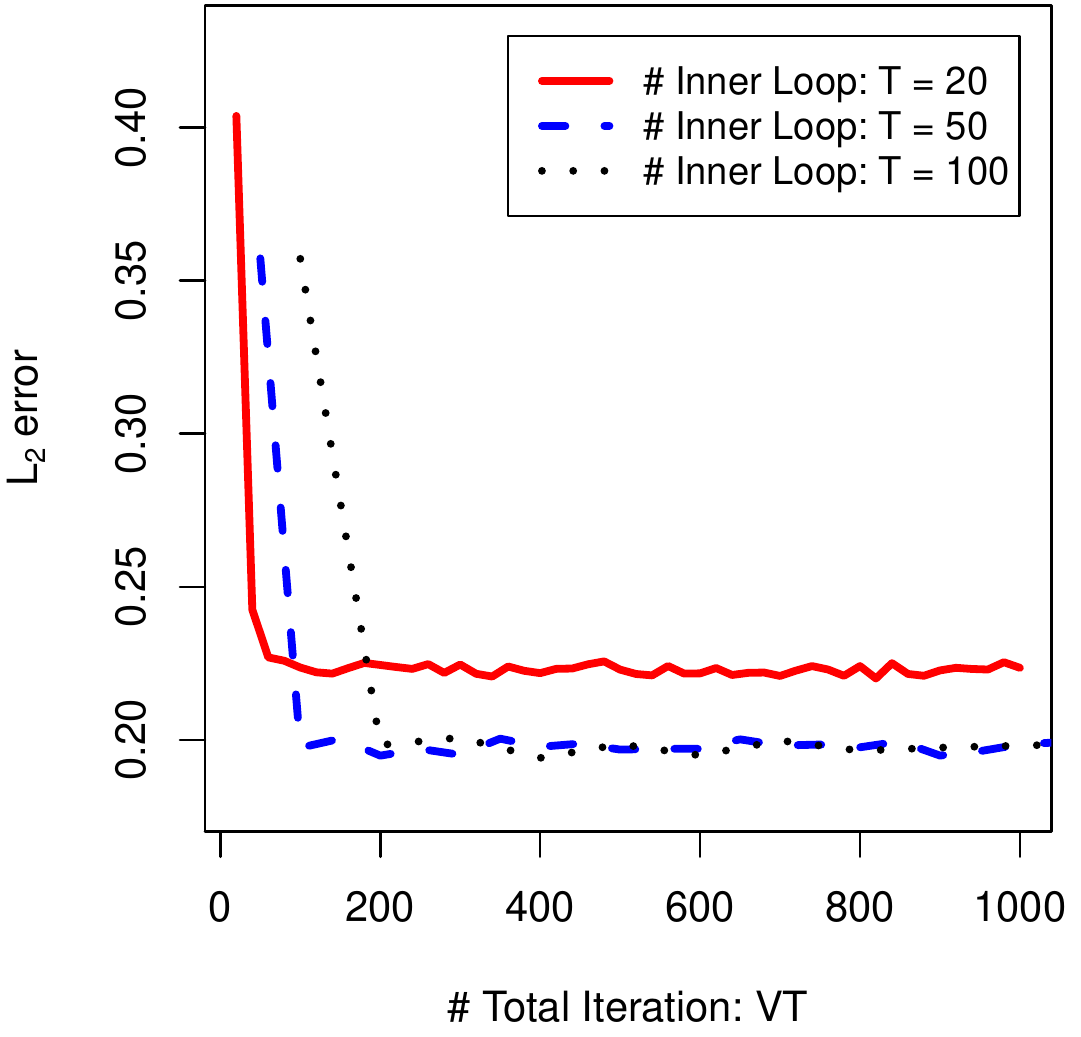}
        \label{fig:simu1_gaussian_total}
    }
    \subfigure[Exponential Noise]{
        \includegraphics[width=0.31\textwidth]{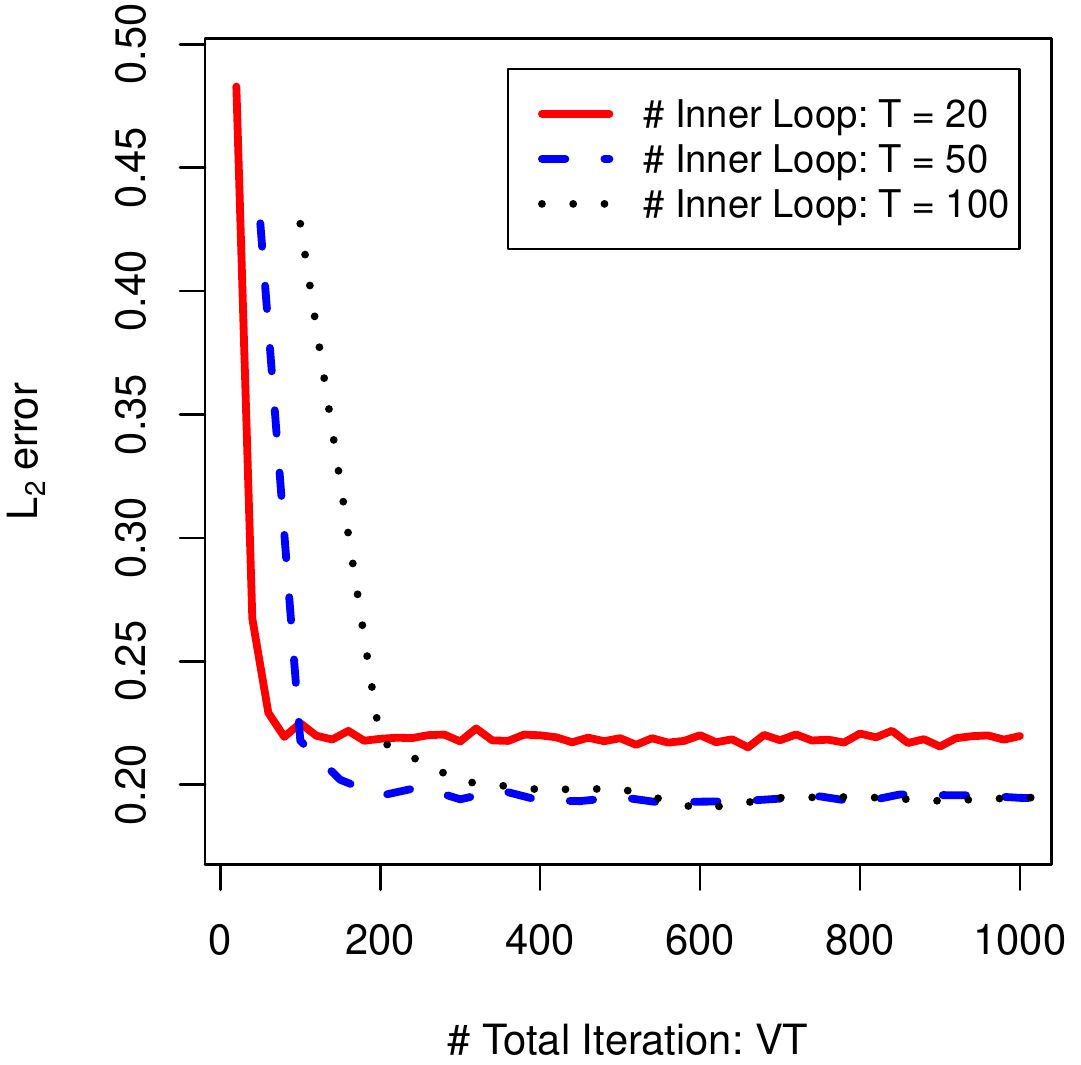}
        \label{fig:simu1_exp_total}
    }
    \subfigure[Cauchy Noise]{
        \includegraphics[width=0.31\textwidth]{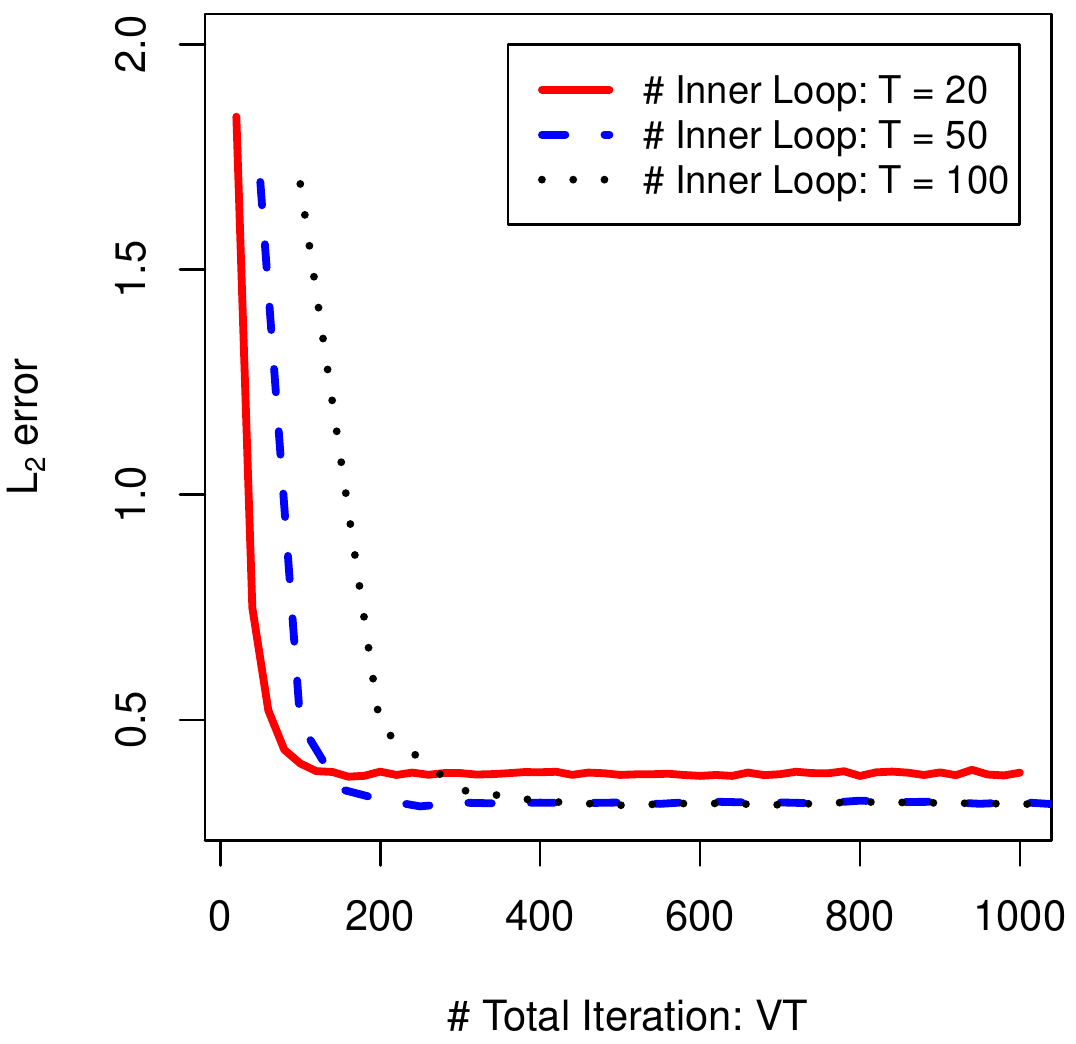}
        \label{fig:simu1_cauchy_total}
    }
      \vspace{-.1in}
    \caption{The $\ell_2$-error from the true QR coefficient versus the number of outer loop iterations and total iterations under different inner loop iterations.}\label{Fig: simu1_effect_of_iteration}
    \vspace{-.1in}
\end{figure*}

In this simulation, we fix the coefficient dimension $p=100$ and local data size $n=200$.
Three different kinds of noises are considered in our simulation, $\text{Normal} (0,1)$, $\text{Exp}(1)$ and $\text{Cauchy}(0,1)$.
We note that the Cauchy distribution belongs to the heavy tail distribution, and its variance is infinite.
We compare the convergence performance of our deSMR method with different inner loop iterations, $T= \{20,50,100\}$.
We evaluate the estimation accuracy by $\ell_2\text{-error}=\sum_{j\in[m]}|\vbeta^{(j)}_{v,T+1} - \vbeta^*_{v}|^2_2$ over outer loop iteration $v = 1,\cdots, 50$. The results are shown in Figure~\ref{Fig: simu1_effect_of_iteration}.

From Figure \ref{fig:simu1_gaussian}-\ref{fig:simu1_cauchy}, it can be seen that our deSMR method converges within $10$ outer loop rounds. 
This matches with our theoretical result in Theorem \ref{thm:betainfV} that the outer loop rounds only need to be of a constant order. 
By comparing the convergence performance over different inner loop iterations, it can be seen that the cases with $T = 50$ and $100$ have very similar performance on $\ell_2$-error, which is smaller than the cases with $T=20$.
This is because our algorithm has not fully converged as the inner loop rounds $T=20$, so that the third term $O_{\mathbb{P}}(\gamma^{T})$ in (\ref{eq:bt}) dominated the other error terms;
But for the cases with $T=50$ and $100$, $O_{\mathbb{P}}(\gamma^{T})$ is negligible and thus they share the similar $\ell_2$-error. This also matches with our theoretical result in Corollary \ref{cor:betainf} that the inner loop rounds only need to be of a constant order when $N=O(\exp(n/s^2))$. 

Furthermore, to have a fairly comparison on the overall complexity, Figure \ref{fig:simu1_gaussian_total}-\ref{fig:simu1_cauchy_total} shows the relationship between $\ell_2$-error and total iteration rounds.
It can be seen that in the cases with $T=50$, our deSMR method converges faster and reaches a lower $\ell_2$-error than the other two settings. 
Therefore, for the rest of the numerical experiments, we use $T=50$ as the inner loop rounds and $V=10$ as the outer loop rounds.

\subsection{Effect of Heavy-Tailed Noise}\label{sec3. Effecti of Heavy-tailed Noise}

The motivation for our robust median regression model is heavy-tailed noises, which is prevalent in real-world applications.
In this part of the simulation study, we aim to compare the estimation performance between our median loss and the standard $\ell_2$ loss in the presence of heavy-tailed noises.
Four difference kinds of noise distribution are considered in this simulation, $\text{Normal} (0,1)$, $\text{Exp}(1)$, $\text{Cauchy}(0,1)$ and $\text{t}(1)$.
Note that both $\text{Cauchy}(0,1)$ and $\text{t}(1)$ are heavy-tailed distribution.
The decentralized median loss can be solved with our proposed deSMR . 
While for the decentralized $\ell_2$ loss, we can solve it with the generalized consensus ADMM method proposed in Section \ref{Sec2. inner_loop}.
We denote this method as Decentralized Lasso Regression (deLR) method.
To fairly compare the two methods, we set the same total iteration number as $500$ for both of them.
Furthermore, we adopt the same setting for the network system as in Section \ref{Sec3. Effect of Iteration}.

We compare the two methods in terms of the following performance metrics: 1) the accuracy of model estimation, $\ell_2\text{-error}=\sum_{j\in[m]}|\widehat{\vbeta}^{(j)} - \vbeta^*|^2_2$; 2) the sparsity recall, which measures the proportion of non-zero coefficient elements that are correctly identified; 3) the sparsity precision, which is defined as the proportion of identified non-zero elements that are correct.
Note that the values of the sparsity recall and precision are in the range $[0, 1]$. The closer to 1, the better the estimation is.
The simulation results are reported in Table \ref{tab:heavy_tail}.

From Table \ref{tab:heavy_tail}, we can see that deLR and deSMR have similar estimation performance when the noises are from Normal(0,1) and Exp(1).
However, for Cauchy(0,1) and t(1), which are heavy-tailed distribution, deSMR has a much better estimation performance than deLR: For example, in the cases with noises from Cauchy(0,1), the $\ell_2$-error of deLR is almost 20 times of deSMR, and the precision of deLR is about 0.2 less than deSMR.
We can see a similar pattern in the cases with noises from t(1).
These findings show that by applying $\ell_2$ loss to the data with heavy-tailed noises, we will obtain a less accurate estimated coefficient with more misidentified non-zero elements.
Thus, compared with the deLR method, our deSMR method has more stable estimation performances under different noise types.

\begin{table*}[t]
\caption{Comparison of $\ell_2$ loss and median loss under difference noises}
\label{tab:heavy_tail}
\begin{center}
\begin{tabular}{l | c | c c c | c c c}
\hline
\hline
\multirow{2}{*}{Noise} &\multirow{2}{*}{$(n,p)$} & \multicolumn{3}{c|}{deLR}  & \multicolumn{3}{c}{deSMR}\\
\hhline{~~------}
 &  & $\ell_2$-error & Recall & Precision & $\ell_2$-error & Recall & Precision\\
 \hline
 \multirow{3}{*}{Normal(0,1)} 
 &  (100,100) & {0.190} & 1.00 & 0.797 & 0.292 & 1.00 & {0.940} \\
 &  (200,100) & {0.131} & 1.00 & 0.832 & 0.200 & 1.00 & {0.951} \\
 &  (200,200) & {0.150} & 1.00 & 0.806 & 0.225 & 1.00 & {0.952} \\
 \hline
 \multirow{3}{*}{Exp(1)} 
 &  (100,100) & {0.261} & 1.00 & 0.808 & 0.292 & 1.00 & {0.955} \\
 &  (200,100) & 0.188 & 1.00 & 0.827 & 0.188 & 1.00 & {0.977} \\
 &  (200,200) & {0.211} & 1.00 & 0.818 & 0.212 & 1.00 & {0.968} \\
\hline
 \multirow{3}{*}{{Cauchy(0,1)}} 
 &  (100,100) & 4.73 & 0.884 & 0.816 & {0.509} & {1.00} & {0.992 } \\
 &  (200,100) & 5.75 & 0.836 & 0.846 & {0.310} & {1.00} & {0.999} \\
 &  (200,200) & 6.36 & 0.815 & 0.843  & {0.351} & {1.00} & {0.999} \\
\hline
 \multirow{3}{*}{{t(1)}} 
 &  (100,100) & 4.72 & 0.891 & 0.839 & {0.503} & {1.00} & {0.991} \\
 &  (200,100) & 6.94 & 0.797 & 0.885 & {0.301} & {1.00} & {0.999} \\
 &  (200,200) & 6.82 & 0.796 & 0.838 & {0.373} & {1.00} & {0.997} \\
\hline
\hline
\end{tabular}
\end{center}
\vskip -0.1in
\end{table*}

\subsection{Effect of Data Heterogeneity}\label{Sec3. Data Heterogeneity}

In this section, we use simulations to illustrate the impact of data heterogeneity on the model estimation. 
We consider generating the data heterogeneity in two ways: 
1) Covariate $\x$ on each node are sampled from Normal$(\0,\vSigma)$ with different covariance matrix.
The covariance matrix $\vSigma$ is constructed by $\vSigma_{ij} = \sigma^2\rho ^{|i-j|}$ for $1\leq i,j\leq p$.
And each node randomly sets $\sigma^2$ and $\rho$ from $\{1,3\}$ and $\{0.1,0.3\}$, respectively. 
In this case, we have the noise from Cauchy(0,1).
2) Noises $\epsilon$ follows different distributions across nodes.
We randomly select one noise distribution from Normal(0,1), Exp(1), Cauchy(0,1) and t(1) for each node.
For this case, the covariate is sampled from the same multivariate normal distribution in Section~\ref{Sec3. Setup}.

We focus on five different methods: 
1) Pooled median regression (Pooled MR), in which local data are pooled into one single machine, and the median regression estimation is calculated with all the data;
2) Local median regression (Local MR), which allows nodes to calculate their own median regression estimation with local data;
3) Averaged median regression (Avg. MR), which takes the average of the local median regression estimation;
4) Decentralized subgradient descent (D-subGD) method \citep{Wang-Li17}, in which the nodes collaboratively solve (\ref{Eq: consensus_problem}) with local subgradient descent and network communication;
5) Our proposed decentralized suggerated median regression (deSMR) method.
For both D-subGD and deSMR, we use the estimations of Local MR as the algorithm initialization. 
We compare the estimation performance of these methods by the $\ell_2$-error and $F_1$-score of estimated regression coefficients.
The $F_1$-score is commonly used metric for support recovery and defined as $F_1\text{-score} = 2 \cdot \text{precision}\cdot\text{recall}/(\text{precision}+\text{recall})$. 
The value of $F_1$-score is between $[0,1]$ and $F_1\text{-score}=1$ implies perfect support recovery.

We report the simulation results in Table \ref{tab:mixing_l2}-\ref{tab:mixing_F1}.
Table \ref{tab:mixing_l2} summarizes the $\ell_2$-error of the five median robust regression methods under difference data heterogeneity settings.
It can be seen that Pooled MR method has the smallest $\ell_2$-error.
However, Pooled MR requires to aggregate all the data into one single machine, which is hard to implement in the decentralized network system and might break the potential data privacy policy.
Our deSMR has the same accuracy as Pooled MR and outperforms other three decentralized median regression methods.
In Table \ref{tab:mixing_F1}, we can see that our deSMR has the $F_1$-score very close to $1$, which means our method has an excellent performance on support recovery. However, the other three decentralized median regression methods have $F_1$-score less than $0.5$ when local data size $m$ equals coefficient dimension $p$.
Thus, we conclude that our deSMR has a better and more stable performance than the other decentralized median regression methods in terms of coefficient $\ell_2$-error and support recovery.

\begin{table*}[t]
\caption{Comparison of $\ell_2$-error of decentralized median robust regression methods under difference data heterogeneities.}
\label{tab:mixing_l2}
\begin{center}
\begin{tabular}{l | c | c | c c  c c }
\hline
\hline
Heter. & $(n,p)$  & Pooled MR & Local MR & Avg. MR & D-subGD & deSMR\\
\hline
 \multirow{3}{*}{Covariate} 
 &  (100,100) & 0.318 & 291 & 142 & 117 & {0.360}  \\
 &  (200,100) & 0.199 & 1.24 & 1.10 & 1.11 & {0.202}  \\
 &  (200,200) & 0.223 & 586 & 325 & 273 & {0.396}  \\
 \hline
  \multirow{3}{*}{Noise} 
 &  (100,100) & 0.401 & 194 & 114 & 66.9 & {0.401}  \\
 &  (200,100) & 0.282 & 1.36 & 1.24 & 0.730 & {0.269}  \\
 &  (200,200) & 0.296 & 466 & 340 & 268 & {0.291}  \\
 \hline
\hline
\end{tabular}
\end{center}
\vskip -0.1in
\end{table*}

\begin{table*}[t]
\caption{Comparison of $F_1$-score of decentralized median robust regression methods under difference data heterogeneities.}
\label{tab:mixing_F1}
\begin{center}
\begin{tabular}{l | c | c | c c  c c }
\hline
\hline
Heter & $(n,p)$  & Pooled MR & Local MR & Avg. MR & D-subGD& deSMR\\
\hline
 \multirow{3}{*}{Covariate} 
 &  (100,100) & 0.995 & 0.186 & 0.182 & 0.266 & {0.988}  \\
 &  (200,100) & 0.998 & 0.959 & 0.737 & 0.837 & {0.998}  \\
 &  (200,200) & 0.998  & 0.101  & 0.0952 & 0.186  & {0.993}  \\
 \hline
  \multirow{3}{*}{Noise} 
 &  (100,100) & 0.988 & 0.186 & 0.182 & 0.297 & {0.991}  \\
 &  (200,100) & 0.994 & 0.932 & 0.626 & 0.810 & {0.993}  \\
 &  (200,200) & 0.997 & 0.107 & 0.0952 & 0.461 & {0.997}  \\
 \hline
\hline
\end{tabular}
\end{center}
\vskip -0.1in
\end{table*}

\subsection{Effect of Network Topology}\label{Sec. Effect of network}

In this section, simulations are conducted to show the impact of network topology on our methods' estimation performance. We focus on two factors: node number and network sparsity.

For numerical study on the node number, we generated total $4000$ data and randomly split them into $m$ nodes, where $m \in \{5,10,20\}$. 
To remove the randomness on the network topology, we consider the decentralized system to be fully connected, in which each node is connected with all other nodes. 
We compare our deSMR method with the other three decentralized median regression methods introduced in Section~\ref{Sec3. Data Heterogeneity}.
We report the simulation results in Table~\ref{tab:Node_l2}.
It can be seen from Table~\ref{tab:Node_l2} that for all the four methods, the $\ell_2$-error gets smaller as the node number decreases.
The reason is that, with less nodes, each node is assigned more data, which leads to a better local initial estimation.
Furthermore, our deSMR method will have a better density estimation (\ref{Eq: density estimation}) with more data. 
Compared with Local MR and Avg. MR, we find that our deSMR method reduces the $\ell_2$-error about 50\% under all the settings.
These results show that our deSMR method can gain more statistical efficiency by having nodes collaboratively estimate the median regression coefficient.

\begin{table}[t]
\caption{Comparison of $\ell_2$-error of decentralized median robust regression methods under difference node numbers.}
\label{tab:Node_l2}
\begin{center}
\begin{tabular}{l |  c | c c  c c }
\hline
\hline
Noise & \# Node  &  Local MR & Avg. MR & D-subGD & deSMR\\
\hline
 \multirow{3}{*}{Normal(0,1)} 
 &  5 &  0.424 & 0.402 & 0.284 & {0.160}  \\
 &  10 &  0.595 & 0.553 & 0.368 & {0.188}  \\
 &  20 &  0.841 & 0.764 & 0.390 & {0.375}  \\
 \hline
  \multirow{3}{*}{Exp(1)} 
 &  5 &  0.431 & 0.409 & 0.299 & {0.153}  \\
 &  10 & 0.656 & 0.613 & 0.381 & {0.178}  \\
 &  20 & 0.967 & 0.880 & 0.434 & {0.427}  \\
 \hline
 \multirow{3}{*}{Cauchy(0,1)} 
 &  5 &  0.692 & 0.662 & 0.585 & {0.274}  \\
 &  10 & 1.26 & 1.20 & 0.742 & {0.307}  \\
 &  20 & 2.40 & 2.24 & {0.900} & 0.967  \\
 \hline
 \multirow{3}{*}{t(1)} 
 &  5 &  0.698 & 0.670& 0.579 & {0.286}  \\
 &  10 & 1.29 & 1.23 & 0.733 & {0.321}  \\
 &  20 & 2.54 & 2.37  & 1.07 & {0.954}  \\
 \hline
\hline
\end{tabular}
\end{center}
\vskip -0.1in
\end{table}

Now, we study the impact of network sparsity by adjusting the probability of network connection $p_c$ in $\{0.3,0.5,0.8\}$.
We fix the total node number as $m=10$, and set local sample size $n=200$, coefficient dimension $p=50$.
The results are shown in Table \ref{tab:network_prob}.
It can be seen that the estimation performances are quite similar under different network sparsity. 
This is mainly due to the fact that, theoretically, the network sparsity affects the convergence factor $\gamma$, and the error term with $\gamma$ diminishes in a linear rate (ref. to Proposition \ref{prop:linear}) so that it is negligible compared with the other statistical error terms (i.e. $O_{\mathbb{P}}\Big(\!\!\sqrt{\frac{s\log N}{N}}\!+\!\max_j\{a_{N,0,j}\}\sqrt{\frac{s^2\log N}{n}}\Big)$).

\begin{table}[ht!]
\caption{Estimation Performance Comparison of DeSMR under different connection probability $p_c$.}
\label{tab:network_prob}
\begin{center}
\begin{tabular}{l | c | c c c }
\hline
\hline
Noise &{$p_c$} & $\ell_2$-error & Recall & Precision  \\
 \hline
 \multirow{3}{*}{Normal(0,1)} 
 &  0.3 & 0.178   &  1.00  &   0.929 \\
 &  0.5 & 0.176   &  1.00  &   0.934  \\
 &  0.8 & 0.179   &  1.00  &   0.934 \\
 \hline
 \multirow{3}{*}{Exp(1)} 
 &  0.3 & 0.173   &   1.00  &   0.949 \\ 
 &  0.5 & 0.169   &   1.00  &   0.940 \\
 &  0.8 & 0.176   &   1.00  &   0.955 \\
 \hline
 \multirow{3}{*}{Cauchy(0,1)} 
 &  0.3 & 0.303  &   1.00   &  0.995  \\
 &  0.5 & 0.299  &   1.00   &  0.991  \\
 &  0.8 & 0.300  &   1.00   &  0.990 \\
 \hline
 \multirow{3}{*}{t(1)} 
 &  0.3 & 0.297  &   1.00&     0.992  \\
 &  0.5 & 0.296  &   1.00&    0.993  \\
 &  0.8 & 0.302  &   1.00&     0.996 \\
\hline
\hline
\end{tabular}
\end{center}
\vskip -0.1in
\end{table}

\subsection{Sensitivity Study of Initial Estimation Method}\label{Sec. Sensitivity study of initialization}

Our theoretical study shows that the initial estimation is an essential step to guarantee the overall convergence of our proposed deSMR method.
In this section, we give a sensitivity study of different initial values: 1) Lasso $\ell_2$ estimator by solving $\min\limits_{\vbeta\in R^{p}} \frac{1}{n} \Abs{\y^{(j)} -\bX^{(j)\rm T}\vbeta}_2^2 + \lambda_{0,j}\abs{\vbeta}_1$ $\forall j= 1, \cdots, m$; 2) Lasso median estimator as in \eqref{Eq: local QR}; 3) true parameter with normal noise perturbation $\widehat{\vbeta}_0^{(j)} = (\widehat{\vbeta}_{0,1}^{(j)},\cdots, \widehat{\vbeta}_{0,p}^{(j)})$, where $\widehat{\vbeta}_{0,i}^{(j)} = \vbeta_i^*+1\{\vbeta_i^*\neq 0\}\cdot N(0,\sigma^2)$, for all $i = 1, \cdots, p$ and $j = 1,\cdots, m$. 
Here we consider two different scale of noise, $\sigma = 0.1$ and $0.5$.
As the ``bad'' initialization will diverge our algorithm, we only compare the estimation performance after one round outer loop iteration. 
The rest settings are the same as in Section~\ref{Sec3. Effect of Iteration}.

The comparison results are shown in Table~\ref{tab: initial_sensitivity}.
First, by comparing the results from different noise perturbations, we can see that the estimation performance gets better with more precise initial values.
Then, by comparing the results from Lasso $\ell_2$ and Lasso median estimators, we find that when the measurement noise $\epsilon$ is from heavy-tailed distribution, using Lasso $\ell_2$ estimator will lead to a large $\ell_2$-error and diverge the algorithm. 
However, with Lasso median estimator as initial value, the algorithm is stable under different measurement noise $\epsilon$.
Thus, we'd suggest Lasso median estimation method as initial value selection for our deSMR algorithm, especially when the measurement error $\epsilon$ comes from heavy-tailed distributions.

\begin{table*}[ht!]
\caption{Sensitivity study of the initial estimation. The results are based on one round outer loop iteration.}
\label{tab: initial_sensitivity}
\begin{center}
\begin{tabular}{l | c c | c c | c c | c c }
\hline
\hline
Initial Method & \multicolumn{2}{c|}{Normal(0,1)}  & \multicolumn{2}{c|}{Exp(1)} & \multicolumn{2}{c|}{Cauchy(1)} & \multicolumn{2}{c}{t(1)}\\
\hhline{~--------}
  &  $\ell_2$-error & $F_1$ & $\ell_2$-error & $F_1$ & $\ell_2$-error & $F_1$ & $\ell_2$-error & $F_1$\\
 \hline
Lasso $\ell_2$ & 0.204 & 0.974 & 0.215 & 0.975 &  7.62 & 0.653 & 6.61 & 0.665 \\ 
 \hline
Lasso Median & 0.360 & 0.958 & 0.422 & 0.966 & 1.40 & 0.982 & 1.50 & 0.983\\
 \hline
$\vbeta^*$+Normal(0,0.1) & 0.211 & 0.969 & 0.211 & 0.981 & 0.322 & 0.999 & 0.317 & 0.998\\
 \hline
$\vbeta^*$+Normal(0,0.5) & 0.331 & 0.940 & 0.344 & 0.935 & 0.556 & 0.992 & 0.585 & 0.989\\
 \hline
\hline
\end{tabular}
\end{center}
\vskip -0.1in
\end{table*}

\section{Real Data Study}\label{Section: realDataStudy}

In this section, we study the Communities and Crime dataset from the UCI Machine Learning Repository \citep{uciCrimeData}. 
We aim to identify the demographic variables that are significantly related to community crime and analyze their linear relationship.

\textbf{Dataset Description.} 
The Communities and Crime dataset include the socio-economic data from the 1990 US Census, law enforcement data from the 1990 Law Enforcement Management and Administrative Statistics survey, and crime data from the 1995 FBI Uniform Crime Report.
This dataset contains a total number of $147$ variables and $2215$ communities in the $49$ states of the United States.
In our analysis, the response variable is the total number of violent crimes per $100K$ population (ViolentCrimesPerPop).
By removing the missing values and scaling the data, we obtain the final dataset with $101$ variables and $1993$ communities.
We randomly split $80\%$ communities as training data and the other $20\%$ as testing data.

\begin{figure}[!ht]
     \centering
     \includegraphics[width=0.43\textwidth]{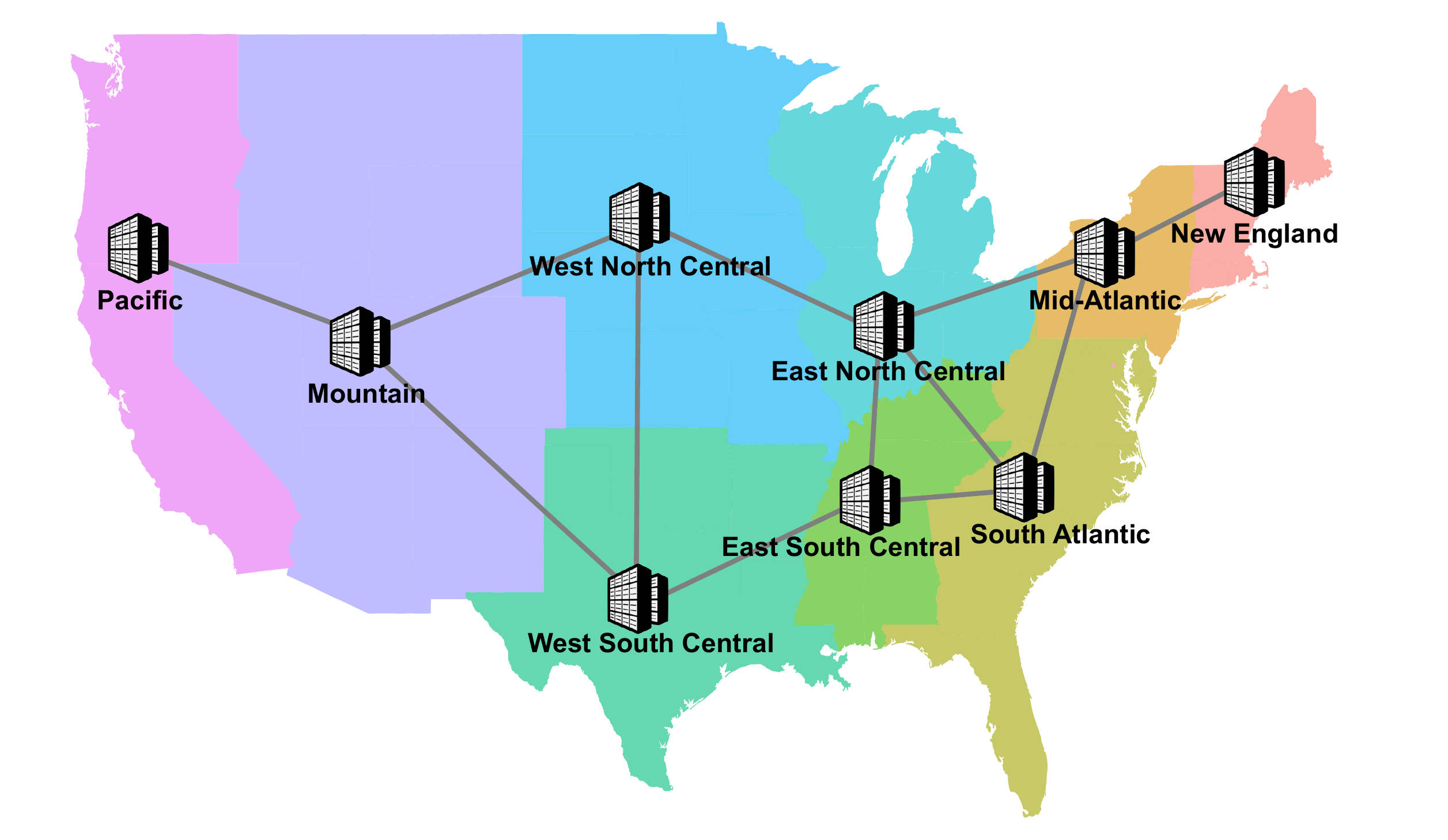}
     \vspace{-.2in}
     \caption{ Decentralized system for UCI communities and crime data study.}\label{Fig: systemDesign_realData}
     \vspace{-.1in}
 \end{figure}

\textbf{System Design.} 
Following the data preprocessing in \cite{yang2019high}, we assign the communities into its Census Bureau-designated division.
There is a total of $9$ divisions, and each division stores its corresponding communities' data.
In the computation system, each division can only broadcast/receive the local information to/from its spatial neighbors, and we show their network relationship in Figure~\ref{Fig: systemDesign_realData}.

\textbf{Outlier Data Models.} 
To study the robustness of different regression methods, we consider three scenarios:
1) Original Data: Fit the model with training data and evaluate the model with testing data; 
2) Balanced Injection: Inject outlier data into the training data of each division, of which the size is $1/9$ of the original local training data size;
3) Attacker Node: Add an attacker node connected with all existing divisions, which contains outlier data with the size of $1:9$ with respect to the total training data.
For outlier data in scenarios 1 and 2, we randomly generate the values of predictive variables from $\text{Normal} (0,1)$, and the values for the response variable are fixed as $12$, which is almost twice the maximum value of the original response variable.

From the above, we can see that our real data network is connected, which satisfies Assumption \ref{ass:network}. Note that Assumptions \ref{ass:subG}-\ref{ass:irres} are some common assumptions on the covariate $\bX$ which are not easy to valid. Assumption \ref{ass:dimen} is the theoretical order of dimensions that still hold for our real data dataset. Our outlier data models indicate the bounded noises which fulfill Assumption \ref{ass:f0}. In our algorithm, we choose the kernel satisfied Assumption \ref{ass:kernel}. As for the Assumption \ref{ass:initial}, according to \cite{Fan-Fan-Barut14}, our initial estimator obtained by solving the local $\ell_1$-regularized median regression problem (\ref{Eq: local QR}) fulfills.

\textbf{Estimation Results.}
We apply our deSMR method to the data and compare it with deLR.
To measure the estimated model performance, we consider two metrics, root mean square error (RMSE) and mean absolute error (MAE), on the testing data.
The results for three scenarios are summarized in Table \ref{tab:uci_results}.
For the scenario with original data, deLR and deSMR have very similar performances that the error differences are within $5\%$.
For the two scenarios with outlier data, our deSMR reduces RMSE about $23\%$ and MAE about $42\%$ compared with deLR.
Additionally, by adding the outlier data, our deSMR only loses $5\%$ RMSE and $10\%$ MAE, while deLR loses $44\%$ RMSE and $89\%$ MAE.
These results show that our deSMR method has robust performance with contaminated data or attacked systems.

\begin{table}[H]
\caption{Comparison of $\ell_2$ loss and median loss for UCI crime dataset}
\label{tab:uci_results}
\begin{center}
\begin{tabular}{l | c  c| c c | c c }
\hline
\hline
\multirow{2}{*}{Metric} & \multicolumn{2}{c|}{Original Data}  & \multicolumn{2}{c|}{Balanced Injection} & \multicolumn{2}{c}{Attacker Node}\\
\hhline{~------}
 &  deSMR & deLR & deSMR & deLR & deSMR & deLR\\
 \hline
RMSE  & 0.639 & {0.610} & {0.672} & 0.887 & {0.674} & 0.871 \\
MAE   & {0.392} & 0.395 & {0.434} & 0.745 & {0.432} & 0.749 \\
\hline
\hline
\end{tabular}
\end{center}
\vskip -0.1in
\end{table}

\section{Conclusion}\label{Section: conclusion}

In this work, we study the problem of robust sparsity learning over decentralized network.
Our goal is to distributively optimize a $\ell_1$ regularized median loss.
To fast solve the `double' non-smooth minimization problem, we proposed the DeMR ADMM method, which enjoys a simple implementation.
We investigated the theoretical properties of both estimation consistency and algorithm convergence. Our theoretical analysis shows that our proposed estimators can achieve a near-optimal rate by ignoring a logarithmic factor.
Furthermore, we conducted thorough numerical experiments to verify our theoretical results, which showed the advantages of our methods in estimation efficiency and robustness.
In our work, we focus on the linear model. 
An interesting future topic is to generalize our framework to a more general class of regression problems, such as robust logistic regression model.

\bibliographystyle{plain}

\bibliography{reference}

\end{document}